\documentclass[twocolumn]{article}

\PassOptionsToPackage{numbers}{natbib}
\usepackage{bosonai_techreport}

\usepackage[utf8]{inputenc}
\usepackage[T1]{fontenc}
\usepackage[scaled=0.92]{inconsolata}
\usepackage[breaklinks=true,colorlinks=true,pageanchor=false]{hyperref}
\hypersetup{
  linkcolor={blue!50!black},
  citecolor={blue!50!black},
  urlcolor={blue!50!black},
}
\usepackage{url}
\usepackage{booktabs}
\usepackage{amsmath}
\usepackage{amsfonts}
\usepackage{nicefrac}
\usepackage{microtype}
\usepackage{xcolor}
\usepackage{colortbl}
\usepackage{graphicx}
\usepackage{multirow}
\usepackage{subcaption}
\usepackage{enumitem}
\usepackage{tcolorbox}
\usepackage{tikz}
\usetikzlibrary{arrows.meta, positioning, fit, calc, decorations.pathreplacing, shadows}
\definecolor{amber}{HTML}{F59E0B}
\definecolor{violet2}{HTML}{7C3AED}
\usepackage{pifont}
\usepackage{pgfplots}
\pgfplotsset{compat=1.18}
\tcbuselibrary{skins, breakable}

\newcommand{\think}{\textsuperscript{\dag}}

\newenvironment{widefig}{\begin{figure*}}{\end{figure*}}

\newenvironment{widetable}{\begin{table*}}{\end{table*}}

\newtcolorbox{promptbox}[1][]{
  colback=gray!5,
  colframe=gray!50,
  fonttitle=\bfseries\small,
  title={#1},
  breakable,
  enhanced,
  left=6pt, right=6pt, top=4pt, bottom=4pt,
  fontupper=\small\ttfamily,
  before upper={\sloppy\setlength{\emergencystretch}{3em}},
}
\newcommand{\pvar}[1]{\textcolor{teal}{\textlangle #1\textrangle}}

\newtcolorbox{dialoguebox}[1][]{
  colback=blue!2,
  colframe=blue!25,
  fonttitle=\bfseries\small,
  title={#1},
  breakable,
  enhanced,
  left=6pt, right=6pt, top=4pt, bottom=4pt,
  fontupper=\small,
}
\newcommand{\passmark}{\textcolor{green!50!black}{\textbf{Pass}}}
\newcommand{\failmark}{\textcolor{red!70!black}{\textbf{Fail}}}

\makeatletter
\renewcommand{\paragraph}{%
  \@startsection{paragraph}{4}{\z@}%
                {0.4ex \@plus 0.2ex \@minus 0.1ex}%
                {-0.8em}%
                {\normalsize\bfseries}%
}
\makeatother

\title{IHBench: Evaluating Post-Interruption Recovery\\in Voice Agents with Structured Workflows}

\author{%
  Ahmad Salimi \hspace{8mm} Wentao Ma \hspace{8mm} Yuzhi Tang \\
  Boson AI \\
  Toronto, ON, Canada \\
  \texttt{\{ahmad,wentao,yuzhi\}@boson.ai} \\
  \And
  Dongming Shen \hspace{8mm} Mu Li \hspace{8mm} Alex Smola\\
  Boson AI \\
  Santa Clara, CA, USA \\
  \texttt{\{dongming,mu,smola\}@boson.ai} \\
}

\renewcommand{\bosonlogo}{boson}

\bloglink{https://www.boson.ai/blog/ihbench}
\codelink{https://github.com/boson-ai/ihbench}
\datalink{https://huggingface.co/datasets/bosonai/ihbench}

\begin{document}

\maketitle

\begin{abstract}
Voice agents deployed in structured workflows (customer service, healthcare scheduling, account management) must handle frequent user interruptions while maintaining progress through multi-step procedures. Existing benchmarks for speech-capable models focus on the \emph{timing} of interruptions: barge-in detection, endpointing, and turn-taking dynamics. They leave unmeasured what happens \emph{after} the interruption: does the agent resume the workflow at the correct step? Does it address the user's interjection? Does it avoid re-delivering content the user already heard?

We introduce \textsc{IHBench} (Interruption Handling Benchmark), a benchmark that evaluates post-interruption recovery in voice agents executing state-machine-driven workflows across 10 enterprise domains. Six interruption types are injected at controlled points mid-utterance, with per-interruption evaluation rubrics generated alongside the data. Each interruption is scored on two axes: \emph{task fulfillment} and \emph{recovery quality}.

We evaluate 27 audio-language model configurations from OpenAI, Google, and the open-weight community. Models vary widely, and recovery quality depends strongly on the interruption type. Across our experiments, closed-weight models are consistently more robust to interruptions than open-weight ones: they win far more often on task fulfillment, degrade roughly $3.3\times$ more slowly as conversations grow longer, and show no audio-versus-text modality gap, whereas the open-weight models lose ground on all three. A human study validates the LLM judge against human annotators, and a cross-benchmark analysis against AudioMultiChallenge indicates that recovery quality is a largely distinct capability axis.
\end{abstract}

\section{Introduction}
\label{sec:intro}

\begin{figure}[t!]
\centering
\includegraphics[width=\columnwidth]{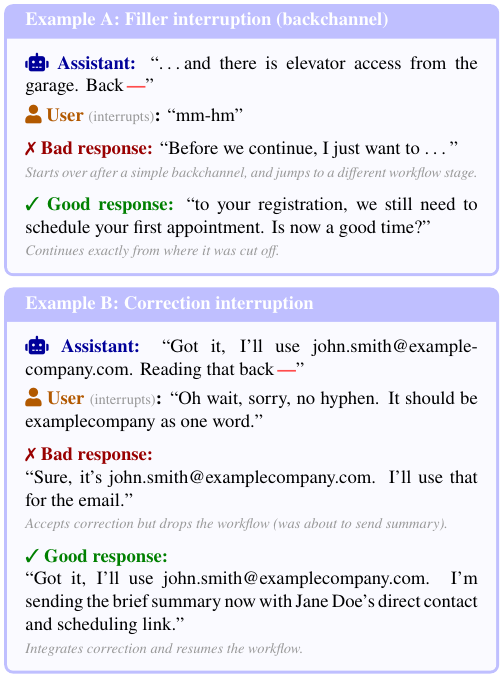}
\vspace{-15pt}
\caption{%
  Good vs.\ bad post-interruption recovery on two \textsc{IHBench} cases: a filler backchannel \textbf{(A)} and a correction \textbf{(B)}. ``\textcolor{red!70}{\textbf{---}}'' marks where the user interrupted mid-utterance.
}
\label{fig:examples}
\vspace{-15pt}
\end{figure}

Real-time voice agents are rapidly moving from research prototypes to production deployments. Models such as GPT Realtime \citep{openai2025gptrealtime}, Gemini \citep{google2025gemini3}, and Moshi \citep{defossez2024moshi} can now process and generate speech natively, enabling voice-first applications in customer service, healthcare, and enterprise workflows. A defining characteristic of natural spoken conversation, one that sharply distinguishes it from text chat, is \emph{interruption}. Users cut in mid-sentence to correct information, express impatience, switch topics, or simply acknowledge with a backchannel like ``mm-hm.'' In structured workflows, these interruptions are not edge cases; they are the norm.

A growing body of work evaluates how well speech models handle the \emph{mechanics} of interruption. Full-Duplex-Bench \citep{cheng2025fullduplexbench} and its successors \citep{cheng2025fullduplexbenchv2, cheng2026fullduplexbenchv3, tang2026instructfd} measure whether models stop speaking when interrupted, how quickly they yield the floor, and whether they distinguish genuine interruptions from backchannels. FLEXI \citep{wang2025flexi} adds a topic-shift score. SID-Bench \citep{liu2026sidbench} focuses on semantic-aware interruption detection. HumDial \citep{chou2026humdial} classifies post-overlap behavior into respond/resume categories.

These benchmarks answer an important question: \emph{can the model detect and react to an interruption in real time?} However, a second question remains largely unanswered, especially in the context of structured task workflows: \emph{what does the model say next?}

Consider a voice agent guiding a user through an insurance claim. The agent is mid-sentence explaining the required documentation when the user cuts in: \emph{``Actually, it was my work address, not home.''} The agent must now (1) stop speaking, (2) recognize this as a correction of previously provided information, (3) integrate the correction, (4) \emph{not} repeat the part of its explanation the user already heard, and (5) resume the workflow at the correct step. Existing benchmarks evaluate step (1). Steps (2)--(5) are the subject of this paper.

The distinction matters because the failure modes are qualitatively different. A model that fails to stop speaking produces an awkward overlap, annoying but recoverable. A model that stops correctly but then re-reads its entire interrupted sentence, ignores the user's correction, or loses its place in the workflow creates a fundamentally broken interaction. Figure~\ref{fig:examples} contrasts good and bad post-interruption recovery on two such cases. In our evaluation, we find that current models, including frontier ones, exhibit gaps in post-interruption recovery that are not captured by existing benchmarks. For example, all evaluated GPT-family audio models continue an utterance correctly after a backchannel less than a third of the time (filler pass rate between 7\% and 31\%), and the Gemini 2.5 family is markedly better (62\%--68\%) although the newer Gemini 3.x line regresses sharply (13\%--32\%). These are not inherently difficult behaviors; rather, they appear to be undertrained, suggesting that targeted evaluation can surface actionable training gaps.

\paragraph{Contributions.} We make the following contributions:

\begin{enumerate}[leftmargin=*, itemsep=-2pt, topsep=0pt]
  \item We define \emph{post-interruption recovery} as a distinct evaluation axis for voice agents, decomposed into six interruption types, and propose a two-axis scoring for it: comparative \emph{task fulfillment} judging and absolute \emph{recovery quality} assessment, both powered by LLM judges with type-specific criteria (Sections~\ref{sec:interruption_types} and~\ref{sec:evaluation}).

  \item We introduce \textsc{IHBench}, a benchmark of synthetically generated multi-turn conversations grounded in state-machine workflows across 10 enterprise domains. Each conversation includes controlled interruptions with per-interruption evaluation rubrics generated alongside the data (Section~\ref{sec:benchmark_design}).

  \item We evaluate 27 audio-language model configurations along several axes (overall scores, per-interruption-type breakdown, degradation with conversation depth, and audio-versus-text input modality), and verify our findings through inter-judge and human-judge agreement studies and a cross-benchmark comparison with AudioMultiChallenge \citep{arora2025audiomc} (Section~\ref{sec:results}).
\end{enumerate}

\section{Related Work}
\label{sec:related}

\paragraph{Full-duplex and turn-taking benchmarks.}
Recent benchmarks evaluate speech-native models on real-time conversational dynamics. Full-Duplex-Bench (FDB) \citep{cheng2025fullduplexbench} evaluates user pause handling, interruption handling, model backchanneling, and smooth turn-taking. For interruption handling, it uses a single-turn audio input and mainly assesses whether the model's response is semantically relevant to the interruption. FDB v1.5 \citep{cheng2025fullduplexbenchv15} expands the test space to additional overlap scenarios, but still uses single-turn audio inputs and evaluates user interruption by whether the model responds to the interruption. FLEXI \citep{wang2025flexi} also uses a single-turn interruption setup and evaluates whether the model shifts topic after a user interruption. SID-Bench \citep{liu2026sidbench} focuses on distinguishing user interruptions from backchannels, HumDial \citep{chou2026humdial} classifies post-overlap behavior as respond or resume, Talking Turns \citep{arora2025talkingturns} benchmarks turn-taking timing using a trained turn-taking model as judge, and INSTRUCT-FD \citep{tang2026instructfd} tests whether a full-duplex system follows explicit turn-taking instructions. These benchmarks characterize whether a model detects, responds to, shifts after, times, or follows instructions about local overlap, but do not evaluate post-interruption recovery within a structured workflow.

\begin{figure*}[t!]
\centering
\includegraphics[width=\textwidth]{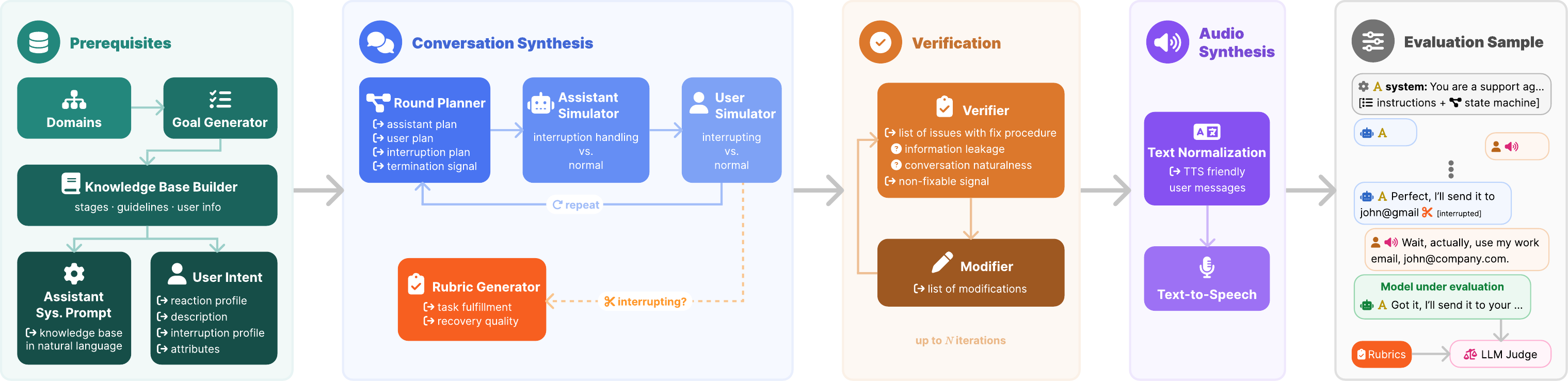}
\caption{%
  The \textsc{IHBench} data generation pipeline: prerequisites, conversation synthesis (with per-interruption rubrics), verification, and audio synthesis. The rightmost panel shows one resulting evaluation sample. Each stage is described in Section~\ref{sec:pipeline}.
}
\label{fig:pipeline}
\vspace{-10pt}
\end{figure*}

FDB v2 \citep{cheng2025fullduplexbenchv2} uses GPT-Realtime \citep{openai2025gptrealtime} as an examiner for multi-turn full-duplex evaluation, and is closest in spirit to our setting. Its interruptions are interjected at pauses surfaced by a voice-activity detector, which keeps the interaction natural but places the cut-in at the model's own turn boundaries rather than at controlled points mid-utterance; the disruptions it studies also center on the correction case. Our pipeline is complementary: because interruptions are scripted into the conversation rather than triggered by a voice-activity detector, they are not confined to detected pauses and can be injected at controlled positions partway through an utterance; it also spans six interruption types beyond correction, with per-interruption rubrics that make the resulting post-interruption recovery directly measurable. MTR-DuplexBench \citep{zhang2026mtrduplexbench} also evaluates multi-turn interruption handling, but focuses on whether the model detects user interruptions and stops speaking in free dialogue. FDB v3 \citep{cheng2026fullduplexbenchv3} evaluates tool-use accuracy under speech disfluencies, but focuses on user self-corrections rather than user-initiated interruptions of the agent. In contrast, our benchmark evaluates whether a voice agent can recover a structured workflow after disruptive mid-speech barge-ins, assessing type-specific recovery behavior.

\paragraph{Voice agents and workflow recovery benchmarks.}
A parallel line of work evaluates task-oriented voice agents on end-to-end task completion: EVA-Bench \citep{chen2026evabench}, VoiceAgentBench \citep{wang2025voiceagentbench}, VoiceBench \citep{chen2024voicebench}, and SpokenWOZ \citep{si2023spokenwoz} measure spoken task success across enterprise domains, tool calling, and dialogue state tracking. $\tau$-voice \citep{sierra2025tauvoice} adds full-duplex turn-taking dynamics through a voice user simulator. None of these, however, inject controlled interruptions or evaluate post-interruption recovery. Closer to interruptions, InterruptBench \citep{gao2026interruptbench} studies mid-task interruptions (additions, revisions, retractions) but on text-based web-navigation agents rather than voice, and work on proactive, transition-aware agents \citep{zeng2025proactive} models text dialogue returning to task after digressions, again without speech-native disruptive interruptions.

\paragraph{Synthetic benchmark construction and evaluation.}
Our data generation pipeline builds on work using LLMs and multi-agent simulation to construct evaluation benchmarks. Self-Instruct \citep{wang2023selfinstruct} and Evol-Instruct \citep{xu2024wizardlm} established iterative LLM generation with quality filtering, while MultiChallenge \citep{sirdeshmukh2025multichallenge} uses a multi-stage multi-agent pipeline with planner, user, and responder agents to generate challenging multi-turn conversations, followed by human review gates and post-hoc binary evaluation questions. SOTOPIA \citep{zhou2023sotopia} similarly simulates multi-agent social interactions with independent agent policies, and Gao et al.\ \citep{gao2026selfevolving} introduce a self-evolving pipeline where judge agents critique intermediate artifacts to refine generation. Our pipeline follows this broader multi-agent design, but adapts it to structured workflow recovery: it generates workflow-grounded conversations with controlled interruption points and fixes each item's per-interruption rubric at construction time (before any model response exists), with a verify--modify loop that enforces state consistency, prevents leakage, and validates recovery targets.

Our evaluation builds on LLM-as-judge methods and rubric-based automatic evaluation, where LLM judges \citep{zheng2023judging, lin2024wildbench} and auto-generated criteria \citep{li2024arenahard} achieve strong agreement with human preferences. We apply this to interruption-specific workflow recovery, using the per-interruption rubrics for both randomized comparative judging and criterion-based absolute recovery evaluation.

\section{Benchmark Design}
\label{sec:benchmark_design}

\textsc{IHBench} evaluates how well voice agents recover after being interrupted during structured, multi-step workflows. The benchmark consists of synthetically generated conversations where an assistant-led state machine drives through an ordered workflow while a user interrupts with realistic spoken-language patterns; Figure~\ref{fig:pipeline} summarizes the full data generation pipeline, which we detail in the rest of this section. We describe the key design choices below.

\subsection{Domains and Workflow Structure}
\label{sec:domains}

Each conversation is grounded in one of 10 enterprise domains (SaaS, financial services, healthcare, telecom, e-commerce, travel, education, government, subscription media, and professional services), selected through an LLM-assisted, manually curated process. For each domain, we use an LLM to generate diverse assistant \emph{goals} (e.g., ``process an insurance claim,'' ``upgrade a subscription plan''), and for each goal, a structured \emph{knowledge base} that defines: \textbf{(i) Ordered workflow stages}: A sequence of states the assistant must traverse (e.g., Authenticate Identity $\rightarrow$ Verify Account $\rightarrow$ Process Request $\rightarrow$ Confirm Changes). Each stage specifies skip conditions, failure handling, and termination conditions. \textbf{(ii) Detailed guidelines}: Domain-specific rules the assistant must follow. \textbf{(iii) Known user information}: Concrete, fictional data values (e.g., account numbers, billing addresses, plan details) that ground the conversation in specific facts rather than placeholders.

The assistant drives the conversation according to a structured workflow and must reach a terminal outcome. This design ensures interruptions are a disruption to \emph{its} plan rather than a natural response to user queries, as would be the case in a free-form conversation with a general-purpose assistant.

\subsection{Interruption Types}
\label{sec:interruption_types}

We define six interruption types, each with distinct recovery requirements:

\noindent\textbf{Normal.} A relevant, cooperative cut-in where the user volunteers a new detail or constraint, or asks a quick clarification (e.g., ``By the way, I'll need this done before Friday''); recovery requires addressing the interjection, then resuming the workflow with a fresh start.\\
\textbf{Impatient.} The user wants to speed up or skip explanations; recovery requires skipping the interrupted content and advancing the workflow without re-explaining what was asked to skip.\\
\textbf{Correction.} The user corrects something \emph{they} previously said in a \emph{prior turn} (e.g., ``Actually wait, use my work email instead''); recovery requires accepting the correction without pushback, integrating the corrected value, and continuing.\\
\textbf{Topic switch.} The user introduces an unrelated request (e.g., ``Oh by the way, can you check my outstanding invoices?''); recovery requires addressing the new topic, then steering back to the original workflow without blending the two.\\
\textbf{Filler.} A brief backchannel that does not change the task (e.g., ``mm-hm,'' ``yeah,'' ``right''); recovery requires \emph{exactly continuing} the interrupted utterance from where it was cut off, without repeating, restarting, or acknowledging the filler.\\
\textbf{Pushback.} The user challenges or resists the assistant's claim or request (e.g., ``I'm not comfortable giving that out over the phone''); recovery requires empathetic de-escalation and offering alternatives.

These types were chosen to span the space of interruption intents observed in real customer-service conversations, from cooperative (normal, filler) through directive (impatient, correction) to adversarial (pushback, topic switch). The distribution of interruption types across the benchmark is reported in Appendix~\ref{app:stats}.

\subsection{Conversation Generation Pipeline}
\label{sec:pipeline}

Conversations are generated through a multi-stage pipeline (Figure~\ref{fig:pipeline}):

\noindent\textbf{Prerequisite generation.} For each (domain, goal) pair, we generate a knowledge base, a system prompt, and multiple user intent profiles. Each user intent includes a behavioral profile (e.g., ``cooperative but rushed''), an interruption probability distribution over the six types, and hidden user information that may surface during the conversation.\\
\textbf{Round-by-round simulation.} A \emph{round planner} decides for each turn what the assistant and user will do, whether an interruption occurs, its type, and where in the assistant's utterance the user cuts in. Separate \emph{assistant simulator} and \emph{user simulator} agents then generate the actual spoken utterances. The user simulator sees the full assistant message but is instructed to produce a truncated version showing the exact cutoff point; it must ensure that its utterance only reacts to the truncated portion (what the user actually heard), preventing information leakage. This constraint is enforced both by prompt design and by the downstream verifier (check~H in Appendix~\ref{app:conv_verifier}).\\
\textbf{Rubric generation.} For each interruption, a rubric generator produces a task fulfillment criterion (what the assistant should do next) and 2--4 recovery quality criteria (type-specific pass/fail checks).\\
\textbf{Verification and modification.} A post-hoc verifier scans every message for naturalness violations, state-machine leakage (user referencing internal workflow stages), content from undelivered portions of interrupted messages, and other issues. A modifier applies minimal targeted edits; conversations that cannot be fixed are discarded.

\subsection{Design Principles}
\label{sec:design_principles}

Several design choices are critical to the benchmark's validity:

\noindent\textbf{No state-machine leakage.} The user does not know the assistant's internal workflow. Each agent in the pipeline is prompted to avoid leaking workflow internals into user utterances, though all such shields rely on the underlying model's instruction-following capability. The final safeguard is the post-hoc verifier (check~G in Appendix~\ref{app:conv_verifier}), which flags conversations where the user references internal stage names, workflow order, or knowledge base details they could not know.\\
\textbf{Concrete values everywhere.} All data in knowledge bases and user intents uses concrete, fictional values, never placeholders. This ensures conversations are grounded in specific facts that can be verified for correctness after interruptions.\\
\textbf{Mid-stage interruptions.} Interruptions are designed to occur \emph{within} workflow stages, not at clean boundaries between them. This makes recovery non-trivial: the assistant must determine where it left off within a stage, not simply start the next one.\\
\textbf{Branched simulators.} The assistant and user simulators use completely separate prompt branches for interruption-handling vs.\ normal turns, preventing cross-contamination of interruption awareness into non-interruption turns.

\subsection{Audio Pipeline}
\label{sec:audio_pipeline}

For audio-input evaluation, each user utterance is text-normalized (expanding numbers, dates, currencies, and abbreviations into spoken form), then synthesized to audio via TTS with Whisper ASR \citep{radford2022whisper} verification and retry logic, using per-conversation voices from Common Voice \citep{ardila2020commonvoice}.

\section{Evaluation Methodology}
\label{sec:evaluation}

Each interruption point becomes one evaluation sample: the model receives the system prompt and the conversation history truncated at the interruption, generates the next response, and is scored on two independent axes by LLM judges (Figure~\ref{fig:pipeline}, rightmost panel).

\label{sec:task_fulfillment}
\noindent\textbf{Task fulfillment (comparative).} We compare the model's response against a baseline response generated by GPT-4o Audio \citep{openai2024gpt4o} from the same conversational context. Following standard win-rate evaluation \citep{li2024arenahard}, a comparative judge receives a single evaluation rubric, describing the next goal the assistant should achieve after this interruption, and the two candidate responses in randomized order; it must pick a winner and name a concrete deficiency in the loser (``more detail'' or ``longer'' are not valid reasons). The metric is the \emph{task fulfillment win rate}: the fraction of interruptions where the model's response is preferred over the baseline. A score of 0.50 indicates parity; scores above 0.50 indicate the model outperforms the baseline on task advancement.

\label{sec:recovery_quality}
\noindent\textbf{Recovery quality (absolute).} The recovery quality judge evaluates a single model response against 2--4 type-specific criteria from the rubric. Each criterion is assessed as met or not met, and the overall verdict is mechanically consistent: all criteria met $\rightarrow$ \passmark, any criterion not met $\rightarrow$ \failmark. We report the \emph{recovery quality pass rate}: the fraction of interruptions where all criteria are met. These criteria are tailored to each interruption type (Section~\ref{sec:interruption_types}); an example rubric is shown in Appendix~\ref{app:example}.

\section{Results}
\label{sec:results}

\begin{table}[t]
\centering
\caption{Overall results on \textsc{IHBench} (audio input, 3 epochs). Baseline: GPT-4o Audio. Judge: GPT-5.4-mini (high reasoning). Mean $\pm$ half-width of 95\% percentile bootstrap CI.}
\label{tab:overall}
\small
\setlength{\tabcolsep}{4pt} %
\begin{tabular}{lcc}
\toprule
\textbf{Model} & \textbf{TF Win Rate} $\uparrow$ & \textbf{RQ Pass Rate} $\uparrow$ \\ %
\midrule
GPT Realtime 2 (medium)\think & \cellcolor{green!45}.728{\tiny$\pm$.03} & .624{\tiny$\pm$.04} \\ %
GPT Realtime 2 (xhigh)\think  & \cellcolor{green!28}.702{\tiny$\pm$.04} & .613{\tiny$\pm$.04} \\ %
GPT Realtime 1.5            & \cellcolor{green!15}.654{\tiny$\pm$.04} & .655{\tiny$\pm$.04} \\ %
GPT Audio                   & .644{\tiny$\pm$.04} & .649{\tiny$\pm$.04} \\ %
Gemini 3 Flash\think        & .632{\tiny$\pm$.03} & .605{\tiny$\pm$.04} \\ %
Gemini 3 Flash              & .598{\tiny$\pm$.04} & .661{\tiny$\pm$.04} \\ %
GPT Realtime                & .597{\tiny$\pm$.04} & \cellcolor{green!15}.680{\tiny$\pm$.04} \\ %
Gemini 2.5 Flash\think      & .586{\tiny$\pm$.04} & \cellcolor{green!45}.704{\tiny$\pm$.04} \\ %
Gemini 3.1 Pro\think        & .582{\tiny$\pm$.04} & .649{\tiny$\pm$.04} \\ %
Gemini 2.5 Pro\think        & .526{\tiny$\pm$.04} & \cellcolor{green!28}.695{\tiny$\pm$.04} \\ %
\rowcolor{blue!7} GPT-4o Audio (baseline) & .500\rlap{$^\ast$} & .654{\tiny$\pm$.05} \\ %
Gemini 2.5 Flash            & .488{\tiny$\pm$.04} & .679{\tiny$\pm$.04} \\ %
GPT Audio Mini              & .484{\tiny$\pm$.04} & .579{\tiny$\pm$.04} \\ %
Gemini 3.1 Flash Live       & .419{\tiny$\pm$.04} & .611{\tiny$\pm$.04} \\ %
GPT Realtime Mini           & .417{\tiny$\pm$.03} & .621{\tiny$\pm$.04} \\ %
Gemini 3.1 Flash Live\think & .405{\tiny$\pm$.04} & .603{\tiny$\pm$.04} \\ %
GPT-4o Mini Audio   & .351{\tiny$\pm$.04} & .654{\tiny$\pm$.05} \\ %
\midrule
Gemma 4 12B Instruct\think  & .511{\tiny$\pm$.03} & .550{\tiny$\pm$.04} \\ %
Gemma 4 12B Instruct        & .505{\tiny$\pm$.04} & .540{\tiny$\pm$.04} \\ %
MiMo-Audio-7B\think         & .445{\tiny$\pm$.04} & .519{\tiny$\pm$.04} \\ %
MiMo-Audio-7B               & .337{\tiny$\pm$.03} & .581{\tiny$\pm$.04} \\ %
Voxtral-Small-24B         & .308{\tiny$\pm$.03} & .593{\tiny$\pm$.04} \\ %
Qwen3-Omni-30B            & .304{\tiny$\pm$.03} & .676{\tiny$\pm$.04} \\ %
Kimi-Audio-7B             & .220{\tiny$\pm$.03} & \cellcolor{red!15}.519{\tiny$\pm$.04} \\ %
Qwen2.5-Omni-7B           & \cellcolor{red!15}.181{\tiny$\pm$.03} & .530{\tiny$\pm$.04} \\ %
Phi-4-Multimodal          & \cellcolor{red!28}.104{\tiny$\pm$.02} & \cellcolor{red!28}.465{\tiny$\pm$.04} \\ %
Qwen2-Audio-7B            & \cellcolor{red!45}.044{\tiny$\pm$.01} & \cellcolor{red!45}.395{\tiny$\pm$.04} \\ %
\bottomrule
\end{tabular}

{\footnotesize $\dag$~Thinking/reasoning-enabled.\quad $^\ast$~TF win rate is measured against the baseline, so TF is $0.50$ by construction for \textit{baseline vs. baseline}.}
\vspace{-15pt}
\end{table}

We evaluate 27 model configurations on \textsc{IHBench} using GPT-5.4-mini \citep{openai2026gpt54mini} (high reasoning) as the judge. The 27 configurations come from 17 closed-weight systems and 10 open-weight configurations. The evaluation is run with three independent epochs per configuration, and 95\% confidence intervals are computed by a 1000-iteration percentile bootstrap~\citep{efron1979bootstrap} over the $N{=}428$ per-sample (epoch-averaged) means.

\subsection{Evaluated Models}
\label{sec:models}

\noindent\textbf{OpenAI models.} GPT-4o Audio and GPT-4o Mini Audio \citep{openai2024gpt4o}; GPT Audio \citep{openai2025gptaudio} and GPT Audio Mini \citep{openai2025gptaudiomini}; GPT Realtime and GPT Realtime 1.5 \citep{openai2025gptrealtime}, GPT Realtime Mini, and GPT Realtime 2 \citep{openai2026gptrealtime2}.\\
\textbf{Google models.} Gemini 2.5 Flash and Gemini 2.5 Pro \citep{google2025gemini25}; Gemini 3 Flash and Gemini 3.1 Pro \citep{google2025gemini3}; and Gemini 3.1 Flash Live \citep{google2026gemini31flashlive}.\\
\textbf{Open-weight models.} Gemma 4 12B Instruct \citep{google2026gemma4}, Qwen3-Omni-30B-A3B-Instruct \citep{xu2025qwen3omni}, Qwen2.5-Omni-7B \citep{xu2025qwen25omni}, Phi-4-Multimodal-Instruct \citep{microsoft2025phi4}, Voxtral-Small-24B-2507 \citep{mistral2025voxtral}, Qwen2-Audio-7B-Instruct \citep{chu2024qwen2audio}, MiMo-Audio-7B-Instruct (in both no-thinking and thinking modes) \citep{xiaomi2025mimoaudio}, and Kimi-Audio-7B-Instruct \citep{moonshot2025kimiaudio}.

\subsection{Overall Results}
\label{sec:overall_results}

Table~\ref{tab:overall} presents the overall results.

\noindent\textbf{Key findings.}
\begin{itemize}[leftmargin=*, itemsep=-2pt, topsep=0pt]
  \item GPT Realtime 2 achieves the highest task fulfillment win rate (0.728), followed by GPT Realtime 2 (xhigh) (0.702) and GPT Realtime 1.5 (0.654).
  \item Recovery quality tells a different story: Gemini 2.5 Flash (thinking) achieves the highest RQ pass rate (0.704) despite a lower TF score (0.586), indicating that task advancement and recovery quality are partially independent axes.
  \item GPT Realtime has the highest RQ among GPT models (0.680) but only mid-tier TF (0.597), suggesting that older Realtime models are more conservative but more correct in recovery behavior.
  \item Thinking/reasoning modes help for Gemini on TF: Gemini 2.5 Flash with thinking outperforms without (0.586 vs. 0.488), and Gemini 3 Flash similarly (0.632 vs. 0.598). However, thinking does not consistently improve RQ.
\end{itemize}

\begin{figure}[t]
\centering
\includegraphics[width=\columnwidth]{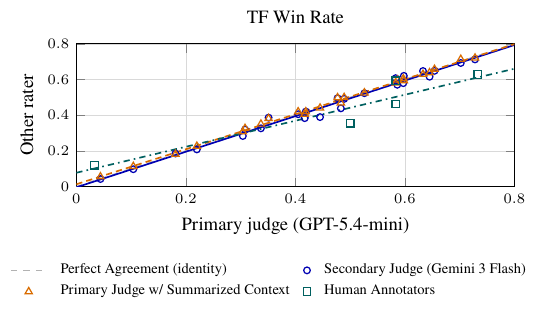}
\vspace{-20pt}
\caption{Per-model task-fulfillment agreement: primary judge (x-axis) vs.\ the secondary judge, the summarized-context judge, and human annotators (y-axis). Colored lines are OLS fits. RQ counterpart: Figure~\ref{fig:inter_judge_rq}.
\vspace{-15pt}}
\label{fig:inter_judge}
\end{figure}

\subsection{Inter-Judge Agreement}
\label{sec:inter_judge}

Every result above rests on a single LLM judge, so we ask whether the rankings are robust to that choice. We validate it two ways: by re-scoring the whole benchmark with an independent second judge from a different provider, and by comparing its verdicts to human annotations. If both judges and human raters recover the same ordering and agree at the verdict level, the conclusions are unlikely to be an artifact of the judging setup.

Figure~\ref{fig:inter_judge} plots the per-model points; the full agreement breakdown across every rater pair on both axes (sample-level agreement, Cohen's $\kappa$, and per-model rank/value correlations) is deferred to Appendix~\ref{app:agreement}, with the headline numbers summarized below.

\begin{widefig}[t]
\centering
\includegraphics[width=\textwidth]{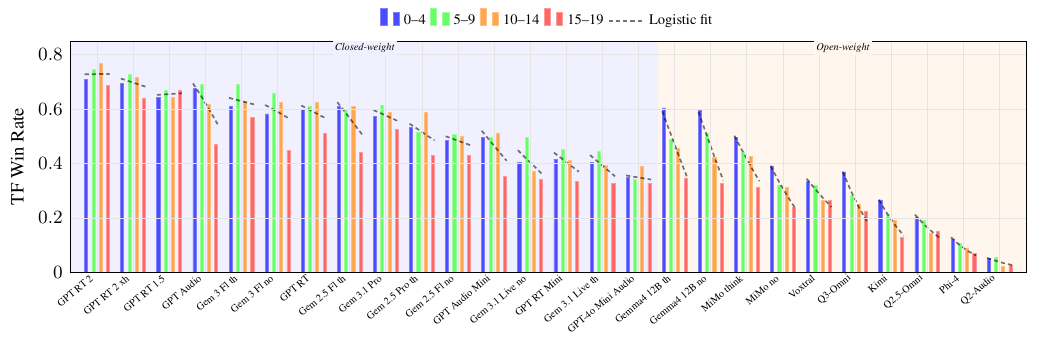}
\vspace{-25pt}
\caption{Task fulfillment win rate by conversation depth, with a per-model logistic regression slope fitted on the raw per-sample data. Pooled across the 26 audio model configurations (excluding TF baseline), the mean slope is significantly negative.}
\label{fig:depth_tf}
\vspace{-10pt}
\end{widefig}

\noindent\textbf{The judges agree with each other.} The second judge is Gemini 3 Flash (with thinking), run over all responses. The model ranking is well preserved (Spearman $\rho = 0.99$ on TF and $0.95$ on RQ) and the two judges' 3-epoch verdicts agree at the sample level in the ``substantial'' range (Cohen's $\kappa = 0.75$ on TF and $0.70$ on RQ), at or above the inter-judge $\kappa$ values typically reported for LLM-as-judge in MT-Bench \citep{zheng2023judging}. The secondary judge is slightly more lenient on RQ, but the offset is near-constant (it preserves the ranking, and the fit slope is close to $1$), so it is inconsequential.\\
\textbf{The judge agrees with humans.} We ran two Prolific studies, one per axis. The full system message and conversation history are long and mostly irrelevant to any single interruption, so we give each annotator a \emph{summarized context}: an LLM-distilled paragraph of the system message and the conversation history relevant to this turn, a self-contained restatement of the criterion, and the last two rounds as audio (full study design in Appendix~\ref{app:human_study}). The validation is strong on two levels. Per item, the judge agrees with humans slightly more often than two humans agree with each other ($\kappa = 0.45$--$0.51$ vs.\ $0.43$ on TF and $0.41$--$0.44$ vs.\ $0.40$ on RQ), so it is statistically indistinguishable from an additional annotator. At the level of conclusions, \textbf{the human ranking matches the judge's almost exactly}: recovery quality is recovered rank for rank (Spearman $\rho = 1.0$), and task fulfillment differs only by a single adjacent swap of two mid-pack models ($\rho = 0.90$), so every comparative claim in this paper would stand unchanged under human scoring. Since humans saw only the summarized context while the judge sees the full context, we re-ran the judge on the same summary (Figure~\ref{fig:inter_judge}); it agrees closely with the full-context judge, so the comparison is not confounded by the judge's context advantage.

\subsection{Task Fulfillment vs.\ Conversation Depth}
\label{sec:depth}

Figure~\ref{fig:depth_tf} shows task fulfillment win rate across four conversation depth bins (by user message index).

\noindent\textbf{Task fulfillment degrades with depth.} For each model except the GPT-4o Audio baseline, we fit a logistic regression of the binary task-fulfillment outcome on conversation depth (user message index). Twenty-four of the 26 remaining audio model configurations have a negative slope. The mean per-model slope is $-0.030$ per additional turn, and a one-sample $t$-test against zero gives $t(25) = -7.04$, $p < 10^{-6}$, confirming that task fulfillment becomes significantly harder as the conversation lengthens. The steepest declines are all on open-weight models; GPT Realtime 2 and GPT Realtime 1.5 are the only two models with non-negative slopes.\\
\textbf{Open-weight models degrade faster with depth.} The ten open-weight configurations all have negative depth slopes between $-0.034$ and $-0.069$ (mean $-0.053$), while the 16 closed-weight configurations have a mean slope of $-0.016$. A Welch $t$-test of closed-weight versus open-weight slopes gives $t = 7.74$, $p < 10^{-6}$. This gap is consistent with open-weight models being trained primarily on short, single-turn or speech-recognition data, leaving them less robust to long multi-turn dialogue context, but our evaluation cannot separately attribute it to training data, model scale, or modality alignment.

\subsection{Recovery Quality Is a New Capability Axis}
\label{sec:benchselect}

\begin{figure}[t]
\centering
\includegraphics[width=\columnwidth]{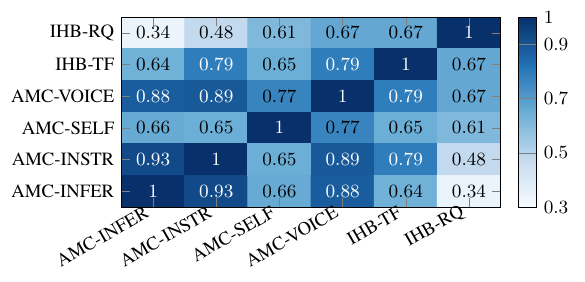}
\vspace{-20pt}
\caption{Pairwise Pearson correlations across the four AMC axes (APR) and the two \textsc{IHBench} primary metrics (TF Win Rate, RQ Pass Rate), over the $n{=}27$ paired models. The IHB-RQ row and column are the lightest in the matrix, reflecting its low correlation with the AMC axes.}
\label{fig:benchselect}
\vspace{-15pt}
\end{figure}

To test whether \textsc{IHBench} adds information beyond existing multi-turn audio evaluation, we pair our 27 model configurations with their per-axis scores on AudioMultiChallenge (AMC)~\citep{arora2025audiomc} and apply the submodular benchmark-selection framework~\citep{smola2026benchselect}. We ran AMC inference ourselves for all 27 configurations (Appendix~\ref{app:amc}). The score matrix has $n{=}27$ paired models and $k{=}6$ evaluation axes: the four AMC axes and the two \textsc{IHBench} primary metrics (TF Win Rate, RQ Pass Rate); their pairwise correlations are shown in Figure~\ref{fig:benchselect}.

\noindent\textbf{The headline.} The four AMC axes intercorrelate strongly ($r = 0.65$--$0.93$), reflecting a dominant general-capability factor, and IHBench TF sits within that band ($\bar{r} = 0.71$). \textbf{IHBench RQ does not}: its mean intercorrelation ($\bar{r} = 0.56$) is the lowest in the $6{\times}6$ matrix, and the gap to the AMC-axis mean excludes zero under a model-level bootstrap~\citep{efron1979bootstrap}. Greedy entropy selection~\citep{smola2026benchselect} also picks IHB-RQ \textbf{\#2}, though partly due to redundancy among the AMC sub-axes. \textbf{IHBench-RQ measures recovery quality}: a capability axis largely distinct from the long-context coherence skills AMC tests.\\
\textbf{IHBench-TF, by contrast, partially overlaps with AMC.} TF Win Rate's mean correlation ($\bar{r} = 0.71$) sits inside the AMC band. TF is essentially a comparative measure of \emph{task progression}, which a more capable multi-turn audio model will tend to do better at regardless of how the user input is delivered, so a strong correlation with a general multi-turn benchmark like AMC is unsurprising. The residual disagreement is still substantive: GPT Realtime 2 is ranked $\#1$ on TF and $\#4$ on AMC, while Gemini 3.1 Pro is ranked $\#9$ on TF and $\#1$ on AMC, indicating that even within the overlap, the specific demand of resuming a workflow after a mid-utterance interruption pulls model families apart in a way the AMC ranking does not capture.\\
\textbf{Recovery quality varies sharply by interruption type.} Breaking recovery quality down by the six interruption types (full per-model table in Appendix~\ref{app:per_type}) exposes a strong type effect. \emph{Filler} is the single largest model-to-model differentiator: GPT-family models continue an utterance correctly after a backchannel only $7\%$--$31\%$ of the time, the Gemini 2.5 family is markedly better ($62\%$--$68\%$), and Gemini 3.x regresses sharply ($13\%$--$32\%$). \emph{Normal} interruptions ($0.71$--$0.85$ pass) and \emph{topic switches} ($0.65$--$0.90$) are handled well across the board, so the spread is concentrated in the harder types.

\subsection{Text-Only vs. Audio-Input Evaluation}
\label{sec:text_vs_audio}

To measure how much the audio modality itself contributes to the difficulty, we re-ran the six Gemini models and nine of the ten open-weight model configurations using transcripts in place of the audio waveforms (OpenAI models and Kimi-Audio-7B are excluded due to not accepting text-only input). Everything else (conversation history, baseline, judge) is held fixed. Figure~\ref{fig:textonly} compares the two conditions on TF; the recovery-quality counterpart is in Appendix~\ref{app:textonly_rq} (Figure~\ref{fig:textonly_rq}).

\begin{figure}[t]
\centering
\includegraphics[width=\columnwidth]{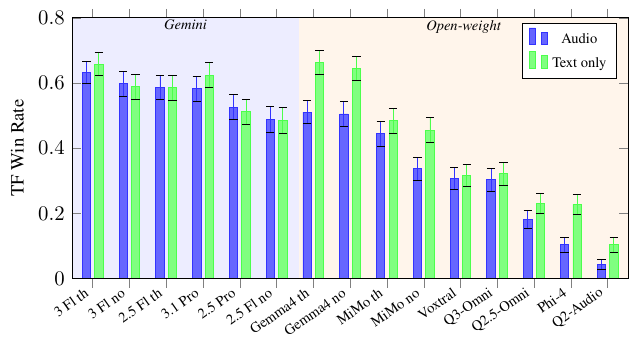}
\vspace{-20pt}
\caption{Audio vs.\ text-only \textbf{task fulfillment} across the 15 dual-modality configurations (6 Gemini + 9 open-weight, 3 epochs each), grouped by family. Error bars are 95\% percentile bootstrap CIs. Audio never beats text-only. The recovery-quality counterpart is in Appendix~\ref{app:textonly_rq}.}
\label{fig:textonly}
\vspace{-10pt}
\end{figure}

\noindent\textbf{Equivalence analysis.} A standard difference test shows only whether modality has \emph{any} effect, not whether the two conditions are comparable, so we use the two one-sided tests (TOST) procedure~\citep{schuirmann1987tost} on per-sample (epoch-averaged) audio$-$text differences, with equivalence margins $\Delta \in \{0.02, 0.03, 0.05\}$. We split by model family, which behave very differently. \\
\textbf{Gemini (six configurations, $n{=}2568$):} both TF (audio$-$text $= -0.007$) and RQ Pass ($-0.001$) are equivalent at $\pm 0.02$ ($p = 0.013$ and $0.0002$, respectively); audio and text-only are statistically indistinguishable.\\
\textbf{Open-weight (nine configurations, $n{=}3852$):} TF audio$-$text $= -0.079$ (text wins by $\sim$8 points) and RQ Pass diff $= -0.062$; TOST fails to establish equivalence even at $\pm 0.05$ on either metric. What is more, a one-sided paired $t$-test shows text significantly outperforms audio on both TF ($t(3851) = -14.95$, $p < 10^{-26}$) and RQ Pass ($t(3851) = -10.70$, $p < 10^{-26}$).\\
\textbf{Audio input never outperforms text.} Across all 15 dual-modality configurations, text-only input is either statistically equivalent to audio (the six Gemini models, within $\pm 0.02$ on every metric) or strictly better (the nine open-weight configurations, text ahead by $\sim$8 points on TF and $\sim$6 on RQ Pass); audio never wins.

\section{Conclusion}
\label{sec:conclusion}

We introduced \textsc{IHBench}, a benchmark for evaluating post-interruption recovery in voice agents executing structured workflows. Unlike existing benchmarks that focus on the timing and mechanics of interruptions, \textsc{IHBench} evaluates what models say \emph{after} an interruption: whether they resume the workflow correctly, address the user's interjection, and avoid common pitfalls like repeating already-delivered content.

Our evaluation of 27 audio-language model configurations (17 closed-weight, 10 open-weight) reveals several findings. The clearest is a consistent gap between closed-weight and open-weight models: the closed-weight models win far more often on task fulfillment, degrade roughly $3.3\times$ more slowly as conversations grow longer, and show no audio-versus-text modality gap, whereas open-weight models lose ground on all three. Beyond this divide:
\begin{itemize}[leftmargin=*, itemsep=-2pt, topsep=0pt]
  \item Post-interruption recovery is a distinct capability with wide spread across models.
  \item Filler handling (resuming an utterance after a backchannel) is a strong family-level differentiator despite being the simplest type conceptually: the GPT family and the newer Gemini 3.x line handle it far less reliably than the Gemini 2.5 family.
  \item Task fulfillment and recovery quality are partially independent axes: the task-fulfillment leader ranks only mid-pack on recovery quality, while the recovery-quality leader has lower task fulfillment.
  \item Against AudioMultiChallenge, IHBench's recovery quality is the lowest-correlated of the six joint evaluation axes (the four AMC axes plus our two metrics). This is evidence that recovery quality is a largely distinct capability axis, not already captured by a strong general audio benchmark.
\end{itemize}

\paragraph{Limitations.}
\textsc{IHBench} is built from synthetic conversations rather than real user interactions, is English-only, and spans 10 enterprise domains; its rubrics inherit the biases of the generator model and the judge. We evaluate recovery on the textual content of responses only, not on prosodic or acoustic recovery behavior (e.g.\ timing or intonation after a barge-in).

\paragraph{Future work.}
Natural extensions include multilingual coverage, validation against real end-user interactions in live deployment, integration with full-duplex timing benchmarks for end-to-end interruption evaluation (timing \emph{and} recovery), and turning the benchmark, or the synthetic pipeline behind it, into a training signal: post-training on its rubrics and generated data to strengthen the recovery behavior we currently only measure.

\bibliographystyle{plain}
\bibliography{references}

\begin{thebibliography}{10}

\bibitem{microsoft2025phi4}
Marah Abdin, Jyoti Aneja, Harkirat Behl, S{\'e}bastien Bubeck, Ronen Eldan,
  Suriya Gunasekar, Michael Harrison, Russell~J. Hewett, Mojan Javaheripi,
  Piero Kauffmann, James~R. Lee, Yin~Tat Lee, Yuanzhi Li, Weishung Liu, Caio
  C.~T. Mendes, Anh Nguyen, Eric Price, Gustavo de~Rosa, Olli Saarikivi, Adil
  Salim, Shital Shah, Xin Wang, Rachel Ward, Yue Wu, Dingli Yu, Cyril Zhang,
  and Yi~Zhang.
\newblock Phi-4 technical report, 2024.

\bibitem{ardila2020commonvoice}
Rosana Ardila, Megan Branson, Kelly Davis, Michael Henretty, Michael Kohler,
  Josh Meyer, Reuben Morais, Lindsay Saunders, Francis~M. Tyers, and Gregor
  Weber.
\newblock Common voice: A massively-multilingual speech corpus, 2020.

\bibitem{arora2025talkingturns}
Siddhant Arora, Zhiyun Lu, Chung-Cheng Chiu, Ruoming Pang, and Shinji Watanabe.
\newblock Talking turns: Benchmarking audio foundation models on turn-taking
  dynamics, 2025.

\bibitem{chen2026evabench}
Tara Bogavelli, Gabrielle~Gauthier Melançon, Katrina Stankiewicz, Oluwanifemi
  Bamgbose, Fanny Riols, Hoang~H. Nguyen, Raghav Mehndiratta, Lindsay~Devon
  Brin, Joseph Marinier, Hari Subramani, Anil Madamala, Sridhar~Krishna Nemala,
  and Srinivas Sunkara.
\newblock Eva-bench: A new end-to-end framework for evaluating voice agents,
  2026.

\bibitem{chen2024voicebench}
Yiming Chen, Xianghu Yue, Chen Zhang, Xiaoxue Gao, Robby~T. Tan, and Haizhou
  Li.
\newblock Voicebench: Benchmarking llm-based voice assistants, 2024.

\bibitem{chu2024qwen2audio}
Yunfei Chu, Jin Xu, Qian Yang, Haojie Wei, Xipin Wei, Zhifang Guo, Yichong
  Leng, Yuanjun Lv, Jinzheng He, Junyang Lin, Chang Zhou, and Jingren Zhou.
\newblock {Qwen2-Audio} technical report, 2024.

\bibitem{defossez2024moshi}
Alexandre Défossez, Laurent Mazaré, Manu Orsini, Amélie Royer, Patrick
  Pérez, Hervé Jégou, Edouard Grave, and Neil Zeghidour.
\newblock Moshi: a speech-text foundation model for real-time dialogue, 2024.

\bibitem{efron1979bootstrap}
Bradley Efron.
\newblock Bootstrap methods: Another look at the jackknife.
\newblock {\em The Annals of Statistics}, 7(1):1--26, 1979.

\bibitem{gao2026selfevolving}
Jiaxuan Gao, Jiaao Chen, Chuyi He, Shusheng Xu, Di~Jin, and Yi~Wu.
\newblock From self-evolving synthetic data to verifiable-reward rl:
  Post-training multi-turn interactive tool-using agents, 2026.

\bibitem{wang2025flexi}
Yuan Ge, Saihan Chen, Jingqi Xiao, Xiaoqian Liu, Tong Xiao, Yan Xiang, Zhengtao
  Yu, and Jingbo Zhu.
\newblock Flexi: Benchmarking full-duplex human-llm speech interaction, 2025.

\bibitem{google2025gemini25}
{Google}.
\newblock Gemini 2.5: Our most intelligent {AI} model.
\newblock
  \url{https://blog.google/innovation-and-ai/models-and-research/google-deepmind/gemini-model-thinking-updates-march-2025/},
  2025.
\newblock Accessed: 2026-06-09.

\bibitem{google2025gemini3}
{Google}.
\newblock {Gemini 3}.
\newblock
  \url{https://blog.google/products-and-platforms/products/gemini/gemini-3/},
  2025.
\newblock Accessed: 2026-06-09.

\bibitem{google2026gemini31flashlive}
{Google}.
\newblock {Gemini 3.1 Flash Live}.
\newblock
  \url{https://blog.google/innovation-and-ai/models-and-research/gemini-models/gemini-3-1-flash-live/},
  2026.
\newblock Accessed: 2026-06-09.

\bibitem{google2026gemma4}
{Google DeepMind}.
\newblock Gemma 4.
\newblock \url{https://deepmind.google/models/gemma/gemma-4/}, 2026.
\newblock Accessed: 2026-06-09.

\bibitem{arora2025audiomc}
Advait Gosai, Tyler Vuong, Utkarsh Tyagi, Steven Li, Wenjia You, Miheer Bavare,
  Arda Uçar, Zhongwang Fang, Brian Jang, Bing Liu, and Yunzhong He.
\newblock Audio multichallenge: A multi-turn evaluation of spoken dialogue
  systems on natural human interaction, 2025.

\bibitem{wang2025voiceagentbench}
Dhruv Jain, Harshit Shukla, Gautam Rajeev, Ashish Kulkarni, Chandra Khatri, and
  Shubham Agarwal.
\newblock Voiceagentbench: Are voice assistants ready for agentic tasks?, 2026.

\bibitem{moonshot2025kimiaudio}
KimiTeam, Ding Ding, Zeqian Ju, Yichong Leng, Songxiang Liu, Tong Liu, Zeyu
  Shang, Kai Shen, Wei Song, Xu~Tan, Heyi Tang, Zhengtao Wang, Chu Wei, Yifei
  Xin, Xinran Xu, Jianwei Yu, Yutao Zhang, Xinyu Zhou, Y.~Charles, Jun Chen,
  Yanru Chen, Yulun Du, Weiran He, Zhenxing Hu, Guokun Lai, Qingcheng Li,
  Yangyang Liu, Weidong Sun, Jianzhou Wang, Yuzhi Wang, Yuefeng Wu, Yuxin Wu,
  Dongchao Yang, Hao Yang, Ying Yang, Zhilin Yang, Aoxiong Yin, Ruibin Yuan,
  Yutong Zhang, and Zaida Zhou.
\newblock Kimi-audio technical report, 2025.

\bibitem{li2024arenahard}
Tianle Li, Wei-Lin Chiang, Evan Frick, Lisa Dunlap, Tianhao Wu, Banghua Zhu,
  Joseph~E. Gonzalez, and Ion Stoica.
\newblock From crowdsourced data to high-quality benchmarks: Arena-hard and
  benchbuilder pipeline, 2024.

\bibitem{lin2024wildbench}
Bill~Yuchen Lin, Yuntian Deng, Khyathi Chandu, Faeze Brahman, Abhilasha
  Ravichander, Valentina Pyatkin, Nouha Dziri, Ronan~Le Bras, and Yejin Choi.
\newblock Wildbench: Benchmarking llms with challenging tasks from real users
  in the wild, 2024.

\bibitem{cheng2026fullduplexbenchv3}
Guan-Ting Lin, Chen Chen, Zhehuai Chen, and Hung yi~Lee.
\newblock Full-duplex-bench-v3: Benchmarking tool use for full-duplex voice
  agents under real-world disfluency, 2026.

\bibitem{cheng2025fullduplexbenchv2}
Guan-Ting Lin, Shih-Yun~Shan Kuan, Jiatong Shi, Kai-Wei Chang, Siddhant Arora,
  Shinji Watanabe, and Hung yi~Lee.
\newblock Full-duplex-bench-v2: A multi-turn evaluation framework for duplex
  dialogue systems with an automated examiner, 2026.

\bibitem{cheng2025fullduplexbenchv15}
Guan-Ting Lin, Shih-Yun~Shan Kuan, Qirui Wang, Jiachen Lian, Tingle Li, Shinji
  Watanabe, and Hung yi~Lee.
\newblock Full-duplex-bench v1.5: Evaluating overlap handling for full-duplex
  speech models, 2026.

\bibitem{cheng2025fullduplexbench}
Guan-Ting Lin, Jiachen Lian, Tingle Li, Qirui Wang, Gopala Anumanchipalli,
  Alexander~H. Liu, and Hung yi~Lee.
\newblock Full-duplex-bench: A benchmark to evaluate full-duplex spoken
  dialogue models on turn-taking capabilities, 2025.

\bibitem{mistral2025voxtral}
{Mistral AI}.
\newblock Voxtral, 2025.

\bibitem{openai2024gpt4o}
{OpenAI}.
\newblock Hello {GPT-4o}.
\newblock \url{https://openai.com/index/hello-gpt-4o/}, 2024.
\newblock Accessed: 2026-06-09.

\bibitem{openai2025gptaudio}
{OpenAI}.
\newblock {gpt-audio}.
\newblock \url{https://developers.openai.com/api/docs/models/gpt-audio}, 2025.
\newblock Accessed: 2026-06-09.

\bibitem{openai2025gptaudiomini}
{OpenAI}.
\newblock {gpt-audio-mini}.
\newblock \url{https://developers.openai.com/api/docs/models/gpt-audio-mini},
  2025.
\newblock Accessed: 2026-06-09.

\bibitem{openai2025gptrealtime}
{OpenAI}.
\newblock Introducing {gpt-realtime} and {Realtime API} updates for production
  voice agents.
\newblock \url{https://openai.com/index/introducing-gpt-realtime/}, 2025.
\newblock Accessed: 2026-06-09.

\bibitem{openai2025o4mini}
{OpenAI}.
\newblock Introducing {OpenAI o3} and {o4-mini}.
\newblock \url{https://openai.com/index/introducing-o3-and-o4-mini/}, 2025.
\newblock Accessed: 2026-06-17.

\bibitem{openai2026gptrealtime2}
{OpenAI}.
\newblock Advancing voice intelligence with new models in the {API}.
\newblock
  \url{https://openai.com/index/advancing-voice-intelligence-with-new-models-in-the-api/},
  2026.
\newblock Accessed: 2026-06-09.

\bibitem{openai2026gpt54mini}
{OpenAI}.
\newblock Introducing {GPT-5.4 mini} and {nano}.
\newblock \url{https://openai.com/index/introducing-gpt-5-4-mini-and-nano/},
  2026.
\newblock Accessed: 2026-06-09.

\bibitem{radford2022whisper}
Alec Radford, Jong~Wook Kim, Tao Xu, Greg Brockman, Christine McLeavey, and
  Ilya Sutskever.
\newblock Robust speech recognition via large-scale weak supervision, 2022.

\bibitem{sierra2025tauvoice}
Soham Ray, Keshav Dhandhania, Victor Barres, and Karthik Narasimhan.
\newblock $\tau$-voice: Benchmarking full-duplex voice agents on real-world
  domains, 2026.

\bibitem{schuirmann1987tost}
Donald~J. Schuirmann.
\newblock A comparison of the two one-sided tests procedure and the power
  approach for assessing the equivalence of average bioavailability.
\newblock {\em Journal of Pharmacokinetics and Biopharmaceutics},
  15(6):657--680, 1987.

\bibitem{si2023spokenwoz}
Shuzheng Si, Wentao Ma, Haoyu Gao, Yuchuan Wu, Ting-En Lin, Yinpei Dai, Hangyu
  Li, Rui Yan, Fei Huang, and Yongbin Li.
\newblock Spokenwoz: A large-scale speech-text benchmark for spoken
  task-oriented dialogue agents, 2025.

\bibitem{sirdeshmukh2025multichallenge}
Ved Sirdeshmukh, Kaustubh Deshpande, Johannes Mols, Lifeng Jin, Ed-Yeremai
  Cardona, Dean Lee, Jeremy Kritz, Willow Primack, Summer Yue, and Chen Xing.
\newblock Multichallenge: A realistic multi-turn conversation evaluation
  benchmark challenging to frontier llms, 2025.

\bibitem{smola2026benchselect}
Alexander Smola.
\newblock Submodular benchmark selection.
\newblock {\em arXiv preprint arXiv:2605.02209}, 2026.

\bibitem{tang2026instructfd}
Yuzhi Tang, Wentao Ma, Xiling Zhao, Ahmad Salimi, Sepehr~Harfi Moridani,
  Dongming Shen, Jixuan Wang, Abdulrahman Abdulrazzag, Murdock Aubry, Yu-Hua
  Chen, Daniel Lee, Jaewon Lee, Jonah Mackey, Silin Meng, Nicholas Stranges,
  Chenxu Xiong, Hao Yu, Yi~Zhu, Mu~Li, and Alex Smola.
\newblock {INSTRUCT-FD}: Can your full-duplex speech system follow turn-taking
  instructions?, 2026.

\bibitem{chou2026humdial}
Chengyou Wang, Hongfei Xue, Guojian Li, Zhixian Zhao, Shuiyuan Wang, Shuai
  Wang, Xin Xu, Hui Bu, and Lei Xie.
\newblock Full-duplex interaction in spoken dialogue systems: A comprehensive
  study from the icassp 2026 humdial challenge, 2026.

\bibitem{wang2023selfinstruct}
Yizhong Wang, Yeganeh Kordi, Swaroop Mishra, Alisa Liu, Noah~A. Smith, Daniel
  Khashabi, and Hannaneh Hajishirzi.
\newblock Self-instruct: Aligning language models with self-generated
  instructions, 2023.

\bibitem{liu2026sidbench}
Kangxiang Xia, Bingshen Mu, Xian Shi, Jin Xu, and Lei Xie.
\newblock Semantic-aware interruption detection in spoken dialogue systems:
  Benchmark, metric, and model, 2026.

\bibitem{xiaomi2025mimoaudio}
LLM-Core-Team Xiaomi.
\newblock Mimo-audio: Audio language models are few-shot learners, 2025.

\bibitem{xu2024wizardlm}
Can Xu, Qingfeng Sun, Kai Zheng, Xiubo Geng, Pu~Zhao, Jiazhan Feng, Chongyang
  Tao, Qingwei Lin, and Daxin Jiang.
\newblock Wizardlm: Empowering large pre-trained language models to follow
  complex instructions, 2025.

\bibitem{xu2025qwen25omni}
Jin Xu, Zhifang Guo, Jinzheng He, Hangrui Hu, Ting He, Shuai Bai, Keqin Chen,
  Jialin Wang, Yang Fan, Kai Dang, Bin Zhang, Xiong Wang, Yunfei Chu, and
  Junyang Lin.
\newblock {Qwen2.5-Omni} technical report, 2025.

\bibitem{xu2025qwen3omni}
Jin Xu, Zhifang Guo, Hangrui Hu, Yunfei Chu, Xiong Wang, Jinzheng He, Yuxuan
  Wang, Xian Shi, Ting He, Xinfa Zhu, Yuanjun Lv, Yongqi Wang, Dake Guo,
  He~Wang, Linhan Ma, Pei Zhang, Xinyu Zhang, Hongkun Hao, Zishan Guo, Baosong
  Yang, Bin Zhang, Ziyang Ma, Xipin Wei, Shuai Bai, Keqin Chen, Xuejing Liu,
  Peng Wang, Mingkun Yang, Dayiheng Liu, Xingzhang Ren, Bo~Zheng, Rui Men, Fan
  Zhou, Bowen Yu, Jianxin Yang, Le~Yu, Jingren Zhou, and Junyang Lin.
\newblock Qwen3-omni technical report, 2025.

\bibitem{zeng2025proactive}
Yejin Yoon, Yuri Son, Namyoung So, Minseo Kim, Minsoo Cho, Chanhee Park,
  Seungshin Lee, and Taeuk Kim.
\newblock Beyond task-oriented and chitchat dialogues: Proactive and
  transition-aware conversational agents, 2025.

\bibitem{zhang2026mtrduplexbench}
He~Zhang, Wenqian Cui, Haoning Xu, Xiaohui Li, Lei Zhu, Haoli Bai, Shaohua Ma,
  and Irwin King.
\newblock Mtr-duplexbench: Towards a comprehensive evaluation of multi-round
  conversations for full-duplex speech language models, 2026.

\bibitem{zheng2023judging}
Lianmin Zheng, Wei-Lin Chiang, Ying Sheng, Siyuan Zhuang, Zhanghao Wu, Yonghao
  Zhuang, Zi~Lin, Zhuohan Li, Dacheng Li, Eric~P. Xing, Hao Zhang, Joseph~E.
  Gonzalez, and Ion Stoica.
\newblock Judging llm-as-a-judge with mt-bench and chatbot arena, 2023.

\bibitem{zhou2023sotopia}
Xuhui Zhou, Hao Zhu, Leena Mathur, Ruohong Zhang, Haofei Yu, Zhengyang Qi,
  Louis-Philippe Morency, Yonatan Bisk, Daniel Fried, Graham Neubig, and
  Maarten Sap.
\newblock Sotopia: Interactive evaluation for social intelligence in language
  agents, 2024.

\bibitem{gao2026interruptbench}
Henry~Peng Zou, Chunyu Miao, Wei-Chieh Huang, Yankai Chen, Yue Zhou, Hanrong
  Zhang, Yaozu Wu, Liancheng Fang, Zhengyao Gu, Zhen Zhang, Kening Zheng,
  Fangxin Wang, Yi~Nian, Shanghao Li, Wenzhe Fan, Langzhou He, Weizhi Zhang,
  Xue Liu, and Philip~S. Yu.
\newblock When users change their mind: Evaluating interruptible agents in
  long-horizon web navigation, 2026.

\end{thebibliography}

\clearpage
\appendix

\section{Dataset Statistics}
\label{app:stats}

\begin{table}[h]
\centering
\caption{Dataset statistics.}
\label{tab:dataset_stats}
\small
\begin{tabular}{lr}
\toprule
\textbf{Statistic} & \textbf{Value} \\
\midrule
Conversations & 45 \\
Domains & 10 \\
Total interruption points & 428 \\
Avg.\ messages per conversation & 30.1 \\
Min / max messages & 19 / 40 \\
\midrule
\multicolumn{2}{l}{\emph{Interruption type distribution}} \\
\quad Pushback & 105 (24.5\%) \\
\quad Impatient & 84 (19.6\%) \\
\quad Normal & 81 (18.9\%) \\
\quad Topic switch & 73 (17.1\%) \\
\quad Filler & 60 (14.0\%) \\
\quad Correction & 25 (5.8\%) \\
\midrule
\multicolumn{2}{l}{\emph{Interruptions by depth (user turn index)}} \\
\quad 0--4 & 169 (39.5\%) \\
\quad 5--9 & 130 (30.4\%) \\
\quad 10--14 & 80 (18.7\%) \\
\quad 15--19 & 49 (11.4\%) \\
\bottomrule
\end{tabular}
\end{table}

\paragraph{Interruption type distribution.} The uneven type distribution reflects two factors. First, the user intent generator is prompted to prefer challenging interruption scenarios over cooperative ones. This is directly visible in the user intent interruption profiles: pushback is the dominant type (probability $\geq 0.25$) in 26 of 45 conversations, while correction is dominant in only 6. Second, correction is the only type where the actual count (25, 5.8\%) falls well below what the average profile probability (12.3\%) would predict (${\sim}53$ expected). This gap arises because corrections require the user to have provided a specific piece of information in a prior turn that they can then revise; the round planner under-generates corrections when no natural self-correction opportunity exists. All other types track their profile probabilities within 2--5 percentage points.

\paragraph{Depth distribution.} Interruptions are concentrated in earlier turns (40\% in turns 0--4) and taper off at depth (11\% in turns 15--19). This is a natural consequence of the conversation length distribution: fewer conversations reach deeper turns, so there are fewer interruption opportunities at greater depth.

Each conversation is seeded from a (domain, goal, user intent) triple. Ten domains are covered: SaaS, financial services, healthcare, telecom, e-commerce, travel, education, government, subscription media, and professional services. Five goals are generated per domain, and ten user intents per goal, yielding 500 possible configurations. Fifty are randomly sampled for conversation synthesis; after the verify--modify loop discards unfixable samples and we keep only interruption points that yield a well-formed evaluation item for every model, 45 conversations with 428 interruption points remain. Multiple conversations may share the same (domain, goal) pair but differ in user intent (personality, interruption profile, hidden information).

\section{Full Inter-Rater Agreement Breakdown}
\label{app:agreement}

Table~\ref{tab:agreement} gives the complete agreement breakdown referenced in Section~\ref{sec:inter_judge}: for every rater pair and both axes, the sample-level agreement and Cohen's $\kappa$, and the per-model Spearman $\rho$ (rank) and Pearson $r$ (value) correlations. The headline numbers (judge--judge $\kappa = 0.75$/$0.70$ on TF/RQ; the judge agreeing with humans at least as well as humans agree with each other) are discussed in the main text.

\begin{widetable}[t]
\centering
\caption{Agreement across raters on \textsc{IHBench}, on both axes. \emph{Sample level}: fraction of items where two raters make the same binary call, with Cohen's $\kappa$. \emph{Per model}: Spearman $\rho$ (rank) and Pearson $r$ (value) over the per-model rates. Judge--judge rows use all 25 configurations and the 3-epoch judges; human rows use the 5 study models, the single-epoch primary judge (for parity with single-rating humans), and $n{=}616$ paired decisions per axis. The judge agrees with humans at least as well as two humans agree with each other (last row), on both axes and with either judge.}
\label{tab:agreement}
\small
\setlength{\tabcolsep}{5pt}
\begin{tabular}{llcccccc}
\toprule
& & \multicolumn{3}{c}{\textbf{Task Fulfillment}} & \multicolumn{3}{c}{\textbf{Recovery Quality}} \\
\cmidrule(lr){3-5}\cmidrule(lr){6-8}
\textbf{Rater pair} & \textbf{Scope} & agree\,$\uparrow$ & $\kappa$\,$\uparrow$ & $\rho$\,/\,$r$\,$\uparrow$ & agree\,$\uparrow$ & $\kappa$\,$\uparrow$ & $\rho$\,/\,$r$\,$\uparrow$ \\
\midrule
Primary $\leftrightarrow$ Secondary judge & 25 cfg & .87 & .75 & .99\,/\,.99 & .86 & .70 & .95\,/\,.98 \\
Primary judge $\leftrightarrow$ Human       & 5 cfg & .75 & .51 & .90\,/\,.95 & .71 & .41 & 1.0\,/\,.93 \\
Secondary judge $\leftrightarrow$ Human     & 5 cfg & .73 & .45 & 1.0\,/\,.99 & .74 & .44 & .90\,/\,.91 \\
Human $\leftrightarrow$ Human (inter-annot.) & 5 cfg & .72 & .43 & --- & .73 & .40 & --- \\
\bottomrule
\end{tabular}

\vspace{2pt}
{\footnotesize ``---'' denotes a metric that does not apply: $\rho$/$r$ require two per-model rankings, which the pooled inter-annotator agreement does not define.}
\end{widetable}

\begin{figure}[t]
\centering
\includegraphics[width=\columnwidth]{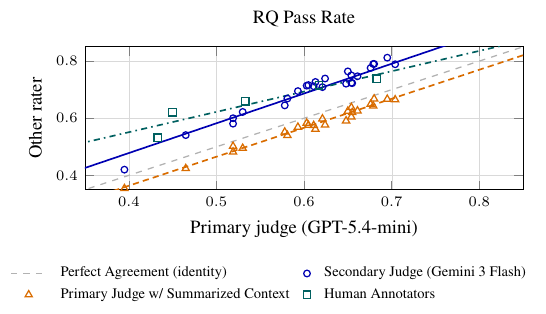}
\caption{Recovery-quality agreement between the primary judge (GPT-5.4-mini, high reasoning, full context; x-axis) and three other raters, per model (the recovery-quality counterpart of Figure~\ref{fig:inter_judge}): \textbf{blue circles} the secondary judge (Gemini 3 Flash with thinking, 25 configurations); \textbf{orange triangles} the same primary judge on the summarized human-study context; \textbf{teal squares} human annotators on the 5 study models (single-epoch primary judge). The dashed gray line is perfect agreement; colored lines are OLS fits. The secondary judge's RQ leniency shows as an offset above the diagonal. Numeric agreement, $\kappa$, and correlations are in Table~\ref{tab:agreement}.}
\label{fig:inter_judge_rq}
\end{figure}

Figure~\ref{fig:inter_judge_rq} is the recovery-quality counterpart of the task-fulfillment agreement plot in the main text (Figure~\ref{fig:inter_judge}), showing the per-model agreement of the primary judge with the secondary judge, the summarized-context judge, and human annotators on the RQ Pass axis. The secondary judge is systematically more lenient than the primary on recovery quality: a per-model OLS fit of secondary on primary RQ Pass rate has slope $1.04$ and intercept $0.06$, so its pass rates sit slightly above the diagonal across models. Because this is a near-constant offset (it preserves the ordering, $\rho = 0.95$ on RQ) and every comparison in the paper uses a single fixed judge, it does not affect any conclusion.

\section{Human Study Design}
\label{app:human_study}

We ran two human studies, one per axis, with annotators recruited through Prolific. Rather than the full system message and conversation history the judge sees, each annotator is shown a \emph{summarized context}: an LLM distills the system message and the parts of the conversation history relevant to the current turn into a single self-contained paragraph, restates the evaluation criterion as a self-contained question, and pairs it with the last two rounds of the conversation rendered as audio. In the task fulfillment study annotators chose which of two responses better satisfied the restated criterion (exactly the comparative call the judge makes); in the recovery quality study they answered the per-criterion yes/no pass questions for a single response. The two studies drew on disjoint annotator pools and each collected $616$ paired human--judge decisions across the 5 study models, with all $300$ items rated by two or more annotators to estimate inter-annotator agreement. The items are a stratified subset of 60 of the 428 interruption points (300 model$\times$sample items per axis), chosen to span interruption types and conversation depth; per-model rates therefore carry sampling uncertainty, while the sample-level agreement in Table~\ref{tab:agreement} is computed on the full set of human judgments.

Human--judge agreement slightly \emph{exceeds} inter-annotator agreement ($\kappa = 0.45$--$0.51$ vs.\ $0.43$ on TF; $0.41$--$0.44$ vs.\ $0.40$ on RQ). This is expected rather than surprising: the judge is a single fixed rater applied uniformly to all items, whereas each study spreads its ${\sim}30$ annotators thinly (about $20$ items each), so the inter-annotator number absorbs between-annotator variance across many different people on top of per-rating noise; a given human therefore agrees somewhat more with the one consistent judge than with another arbitrary annotator. This holds whether the primary or the secondary judge is used, so the validation is not specific to one judge. Per model, the human ranking matches the judge's almost exactly across all five study models (with either judge): recovery quality is recovered rank for rank ($\rho = 1.0$), and task fulfillment differs only by a single adjacent swap of \textsc{gpt-realtime-mini} and \textsc{mimo-audio-7b-thinking} ($\rho = 0.90$), two mid-pack models the judge scores within a few points and that the 3-epoch judge likewise keeps essentially tied ($0.489$ vs.\ $0.494$).

To guard against low-quality annotators inflating this picture, we tested each annotator's agreement with the leave-one-out majority of the other annotators on their shared items (one-sided binomial test against the cohort base rate, Holm-corrected); this criterion never references the judge, so it cannot bias the human--judge comparison. In neither study did any annotator disagree with their peers significantly more than chance after correction, so we exclude none.

\section{Per-Type Recovery Quality Breakdown}
\label{app:per_type}

Table~\ref{tab:per_type} gives the full per-model recovery quality pass rate broken down by the six interruption types, summarized in the main text (Section~\ref{sec:overall_results}).

\begin{widetable}[t]
\centering
\caption{Recovery quality pass rate by interruption type (audio input, 3 epochs). Mean $\pm$ half-width of 95\% percentile bootstrap CI (1000 resamples). Best per column in \textbf{bold}.}
\label{tab:per_type}
\small
\begin{tabular}{lcccccc}
\toprule
\textbf{Model} & \textbf{Normal} & \textbf{Impat.} & \textbf{Corr.} & \textbf{Topic} & \textbf{Filler} & \textbf{Push.} \\
\midrule
GPT Realtime 2 (medium)\think & .76{\tiny$\pm$.08} & .52{\tiny$\pm$.09} & .73{\tiny$\pm$.14} & .76{\tiny$\pm$.08} & .19{\tiny$\pm$.08} & .72{\tiny$\pm$.07} \\
GPT Realtime 2 (xhigh)\think & .78{\tiny$\pm$.08} & .49{\tiny$\pm$.09} & .68{\tiny$\pm$.15} & .76{\tiny$\pm$.08} & .16{\tiny$\pm$.07} & .72{\tiny$\pm$.08} \\
GPT Realtime 1.5            & .82{\tiny$\pm$.07} & .57{\tiny$\pm$.09} & .68{\tiny$\pm$.15} & \textbf{.90}{\tiny$\pm$.06} & .13{\tiny$\pm$.07} & .72{\tiny$\pm$.07} \\
GPT Audio                   & .82{\tiny$\pm$.08} & .54{\tiny$\pm$.10} & \textbf{.76}{\tiny$\pm$.13} & .77{\tiny$\pm$.08} & .22{\tiny$\pm$.09} & .74{\tiny$\pm$.07} \\
Gem.\ 3 Flash\think         & .77{\tiny$\pm$.08} & .54{\tiny$\pm$.10} & .63{\tiny$\pm$.16} & .80{\tiny$\pm$.07} & .14{\tiny$\pm$.07} & .66{\tiny$\pm$.07} \\
Gem.\ 3 Flash               & .81{\tiny$\pm$.07} & .53{\tiny$\pm$.09} & \textbf{.76}{\tiny$\pm$.13} & .87{\tiny$\pm$.06} & .32{\tiny$\pm$.10} & .68{\tiny$\pm$.08} \\
GPT Realtime                & \textbf{.83}{\tiny$\pm$.07} & .60{\tiny$\pm$.09} & .72{\tiny$\pm$.15} & .80{\tiny$\pm$.07} & .31{\tiny$\pm$.11} & .74{\tiny$\pm$.08} \\
Gem.\ 2.5 Flash\think       & .75{\tiny$\pm$.08} & .58{\tiny$\pm$.09} & .63{\tiny$\pm$.15} & .78{\tiny$\pm$.08} & \textbf{.68}{\tiny$\pm$.11} & \textbf{.75}{\tiny$\pm$.08} \\
Gem.\ 3.1 Pro\think         & .82{\tiny$\pm$.07} & .66{\tiny$\pm$.10} & .57{\tiny$\pm$.18} & .82{\tiny$\pm$.08} & .13{\tiny$\pm$.07} & .71{\tiny$\pm$.07} \\
Gem.\ 2.5 Pro\think         & .75{\tiny$\pm$.08} & \textbf{.69}{\tiny$\pm$.08} & .65{\tiny$\pm$.15} & .77{\tiny$\pm$.07} & .62{\tiny$\pm$.10} & .66{\tiny$\pm$.08} \\
Gem.\ 2.5 Flash             & .76{\tiny$\pm$.08} & .58{\tiny$\pm$.09} & .61{\tiny$\pm$.17} & .77{\tiny$\pm$.08} & .64{\tiny$\pm$.10} & .67{\tiny$\pm$.08} \\
GPT Audio Mini              & .77{\tiny$\pm$.08} & .48{\tiny$\pm$.09} & .72{\tiny$\pm$.15} & .65{\tiny$\pm$.08} & .08{\tiny$\pm$.06} & .71{\tiny$\pm$.07} \\
GPT-4o Audio        & .82{\tiny$\pm$.08} & .64{\tiny$\pm$.11} & \textbf{.76}{\tiny$\pm$.16} & .85{\tiny$\pm$.09} & .08{\tiny$\pm$.08} & .71{\tiny$\pm$.09} \\
Gem.\ 3.1 Flash Live        & .75{\tiny$\pm$.08} & .58{\tiny$\pm$.09} & .57{\tiny$\pm$.19} & .74{\tiny$\pm$.08} & .24{\tiny$\pm$.09} & .66{\tiny$\pm$.08} \\
GPT Realtime Mini           & .80{\tiny$\pm$.07} & .55{\tiny$\pm$.09} & .60{\tiny$\pm$.17} & .82{\tiny$\pm$.07} & .07{\tiny$\pm$.06} & .72{\tiny$\pm$.08} \\
Gem.\ 3.1 Flash Live\think  & .74{\tiny$\pm$.08} & .57{\tiny$\pm$.09} & .63{\tiny$\pm$.15} & .74{\tiny$\pm$.08} & .23{\tiny$\pm$.08} & .64{\tiny$\pm$.08} \\
GPT-4o Mini Audio   & \textbf{.83}{\tiny$\pm$.08} & .64{\tiny$\pm$.10} & .52{\tiny$\pm$.20} & .88{\tiny$\pm$.07} & .15{\tiny$\pm$.08} & .70{\tiny$\pm$.09} \\
\midrule
Gemma 4 12B Instruct\think & .66{\tiny$\pm$.09} & .52{\tiny$\pm$.09} & .44{\tiny$\pm$.16} & .70{\tiny$\pm$.07} & .25{\tiny$\pm$.08} & .58{\tiny$\pm$.07} \\
Gemma 4 12B Instruct       & .63{\tiny$\pm$.09} & .51{\tiny$\pm$.09} & .48{\tiny$\pm$.15} & .67{\tiny$\pm$.08} & .22{\tiny$\pm$.08} & .60{\tiny$\pm$.08} \\
MiMo-Audio-7B\think        & .68{\tiny$\pm$.08} & .36{\tiny$\pm$.08} & .47{\tiny$\pm$.16} & .72{\tiny$\pm$.07} & .03{\tiny$\pm$.03} & .68{\tiny$\pm$.07} \\
MiMo-Audio-7B              & .68{\tiny$\pm$.06} & .47{\tiny$\pm$.09} & .45{\tiny$\pm$.15} & .66{\tiny$\pm$.07} & .40{\tiny$\pm$.11} & .67{\tiny$\pm$.07} \\
Voxtral-Small-24B          & .74{\tiny$\pm$.07} & .53{\tiny$\pm$.09} & .49{\tiny$\pm$.15} & .70{\tiny$\pm$.08} & .36{\tiny$\pm$.10} & .61{\tiny$\pm$.08} \\
Qwen3-Omni-30B             & .80{\tiny$\pm$.07} & .63{\tiny$\pm$.08} & .55{\tiny$\pm$.17} & .80{\tiny$\pm$.06} & .46{\tiny$\pm$.10} & .69{\tiny$\pm$.07} \\
Kimi-Audio-7B              & .61{\tiny$\pm$.09} & .45{\tiny$\pm$.08} & .44{\tiny$\pm$.16} & .68{\tiny$\pm$.08} & .32{\tiny$\pm$.10} & .53{\tiny$\pm$.07} \\
Qwen2.5-Omni-7B            & .65{\tiny$\pm$.08} & .53{\tiny$\pm$.08} & .33{\tiny$\pm$.14} & .66{\tiny$\pm$.08} & .35{\tiny$\pm$.09} & .49{\tiny$\pm$.07} \\
Phi-4-Multimodal           & .52{\tiny$\pm$.09} & .56{\tiny$\pm$.08} & .28{\tiny$\pm$.13} & .63{\tiny$\pm$.07} & .18{\tiny$\pm$.07} & .44{\tiny$\pm$.07} \\
Qwen2-Audio-7B             & .51{\tiny$\pm$.10} & .44{\tiny$\pm$.08} & .31{\tiny$\pm$.13} & .58{\tiny$\pm$.10} & .14{\tiny$\pm$.07} & .30{\tiny$\pm$.07} \\
\bottomrule
\end{tabular}

\vspace{2pt}
{\footnotesize $\dag$~Thinking/reasoning-enabled.}
\end{widetable}

\paragraph{Filler is the hardest type for GPT models.} When a user produces a brief backchannel like ``mm-hm'' mid-utterance, the correct recovery is to continue the utterance from where it was cut off, without repeating or restarting. GPT-family models fail at this consistently: filler pass rates are between 0.07 (GPT Realtime Mini) and 0.31 (GPT Realtime). The failure modes are restarting the utterance, acknowledging the filler (``Glad you're following along!''), or producing an entirely new response. The Gemini 2.5 family is markedly better (Gem.\ 2.5 Pro 0.62, Gem.\ 2.5 Flash thinking 0.68, Gem.\ 2.5 Flash 0.64), but Gemini 3.x regresses sharply (0.13--0.32), suggesting filler handling did not carry over to the newer model line. This is the single largest model-to-model gap among the six interruption types.

\paragraph{Normal and topic switch are generally well-handled.} Across models, normal interruptions are handled at $0.71$--$0.85$ pass rate and topic switches at $0.65$--$0.90$, suggesting that addressing a relevant cut-in and engaging with a new topic are relatively natural capabilities for current LLMs.

\paragraph{Impatient and correction are middle-difficulty types.} Impatient recovery (skipping content the user asked to skip) clusters around $0.45$--$0.68$, with Gemini 3.1 Pro the best at $0.66$ and GPT Audio Mini the worst at $0.47$. Correction recovery (accepting a self-correction without pushback) shows a wider spread of $0.52$--$0.75$, partly driven by smaller per-type sample sizes (Section~\ref{app:stats}).

\section{Audio vs.\ Text-Only Recovery Quality}
\label{app:textonly_rq}

\begin{figure}[t]
\centering
\includegraphics[width=\columnwidth]{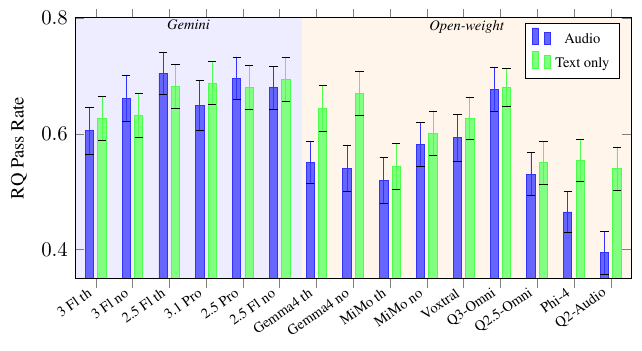}
\caption{Audio vs.\ text-only \textbf{recovery quality} across the 15 dual-modality configurations (6 Gemini + 9 open-weight, 3 epochs each), the recovery-quality counterpart of Figure~\ref{fig:textonly}. Grouping, ordering, family shading, and error bars are as in Figure~\ref{fig:textonly}. As on task fulfillment, Gemini models are within $\pm$0.02 across modalities while the open-weight configurations are better with text input; audio never wins.}
\label{fig:textonly_rq}
\end{figure}

Figure~\ref{fig:textonly_rq} is the recovery-quality counterpart of the task-fulfillment audio-vs-text comparison in the main text (Figure~\ref{fig:textonly}, Section~\ref{sec:text_vs_audio}). The same conclusion holds on the RQ Pass axis: the six frontier Gemini models are statistically equivalent across modalities (within $\pm 0.02$), while the nine open-weight configurations are better with text input (RQ Pass audio$-$text $= -0.062$), and audio never wins. The full TOST equivalence results for both metrics are reported in the main text.

\section{AudioMultiChallenge Scoring}
\label{app:amc}

The cross-benchmark analysis in Section~\ref{sec:benchselect} pairs each of our 27 model configurations with its scores on AudioMultiChallenge (AMC)~\citep{arora2025audiomc}. Several of the evaluated models post-date the AMC paper and are not in its released results, so we ran AMC inference ourselves for all 27 configurations rather than reusing reported numbers, ensuring every model is scored under identical conditions. We use the official AMC judge (\texttt{o4-mini}~\citep{openai2025o4mini}) with the exact judge system prompt and structured-output schema from the AMC dataset card~\citep{arora2025audiomc}, unmodified, and run three epochs per configuration to match our \textsc{IHBench} protocol; the per-axis APR scores used in Figure~\ref{fig:benchselect} are the epoch-averaged means.

\section{Example Conversation with System Prompt and Rubrics}
\label{app:example}

We present a complete example from the benchmark: the assistant's system prompt, the first eight messages of the conversation, and the evaluation rubric for the first interruption. This conversation is from the Government and Public Services domain (disaster housing assistance workflow) and contains 9 interruptions across 5 types: pushback, impatient, topic switch, normal, and correction.

\subsection{Assistant System Prompt}

The system prompt below is generated by the pipeline's system message writer (Appendix~\ref{app:system_message_writer}) from the domain, goal, and knowledge base. It serves as the assistant's complete operating instructions during the conversation.

\begin{promptbox}[System prompt: Disaster Housing Assistance]
\small
You are the call-based assistant for Government and Public Services, supporting public service agencies that manage structured applications, identity verification, compliance steps, and formal registrations. You are representing the State Disaster Housing Assistance (SDHA) program. Your purpose is to proactively reach out to a disaster assistance applicant after initial intake, coordinate a required home inspection, verify case ID and identity (name and DOB), confirm the damaged property address and accessibility constraints, capture preferred inspection windows and any special access needs, obtain explicit consent to share contact details with the inspector and to send a confirmation, and finalize by booking the inspection and providing the appointment confirmation number.

\textbf{How to operate}

- Initiate the conversation. Establish who you are (SDHA), why you're contacting the applicant (inspection coordination for their disaster assistance application), and what will happen in this call (verification, address confirmation, access/safety details, scheduling, and consent). Do not assume the user remembers prior steps.

- Drive a strict workflow that follows the stages below in order. Apply skip conditions only when explicitly met during this same conversation. Do not narrate stage numbers to the user.

- Only use two sources of facts: (1) known\_user\_information (listed below) and (2) information the user provides or confirms in this conversation. Do not invent any other data.

- Ask concise, single questions where possible. Before any irreversible action (capturing consent, booking), read back critical details and confirm.

- Collect only the following data elements: case ID, full name, date of birth, damaged property address, access/safety notes, availability windows, consent to share contact details with the inspector, and consent to send confirmation via their chosen method. Do not request SSN or financial information.

- Handle interruptions: pause to address the concern, then resume from the current stage. If the user wants to stop, confirm and terminate respectfully.

\ldots\ \emph{[additional operating rules: privacy protection rationale, distress handling, summary requirements, inspection window policy, safety/access disclosure, communication consent, data minimization]}

\textbf{Known user information} (use only as facts; do not alter unless the user corrects them)

\textbf{program\_name:} State Disaster Housing Assistance (SDHA)\\
\textbf{case\_id:} DR-4827315\\
\textbf{applicant\_full\_name:} Marisol Ortega\\
\textbf{applicant\_dob:} 1984-11-02\\
\textbf{damaged\_property\_address:} 7420 Magnolia Ridge Dr, Lakeview, LA 70121\\
\textbf{mailing\_address\_on\_file:} PO Box 1184, Metairie, LA 70001\\
\textbf{primary\_phone\_on\_file:} 504-555-2974\\
\textbf{secondary\_phone\_on\_file:} 985-555-8039\\
\textbf{email\_on\_file:} marisol.ortega84@gmail.com\\
\textbf{language\_preference:} English\\
\textbf{applicant\_time\_zone:} Central Time (CT)\\
\textbf{intake\_submission\_date:} 2026-02-17\\
\textbf{initial\_accessibility\_notes:} Driveway partially blocked by a downed oak; small dog named Coco; electrical power restored on 2026-02-28\\
\textbf{inspection\_window\_policy:} Weekdays Mon--Fri: AM 8--12, PM 12--4, Late 4--6; Limited Saturday AM slots\\
\textbf{presence\_requirement:} Applicant or another adult (18+) with photo ID must be present for the inspection\\
\textbf{safety\_preparation\_requirements:} Secure pets, clear pathways to all rooms and utilities, disclose any hazards (mold, gas leaks, structural instability)\\
\textbf{consent\_to\_share\_contact\_with\_inspector\_status:} Not yet obtained\\
\textbf{earliest\_available\_inspection\_date:} 2026-03-09\\
\textbf{inspection\_estimated\_duration:} 60--90 minutes\\
\textbf{booking\_hold\_limit\_minutes:} 30\\
\textbf{inspection\_appointment\_confirmation\_number:} INS-20260309-18427\\
\textbf{conversation\_reference\_id:} CR-20260305-6102

\textbf{State machine: required stages and controls}

\textbf{1) Open and Establish Purpose}\\
Introduce yourself as calling from SDHA about the applicant's disaster assistance application to coordinate a required home inspection. Confirm you're speaking with the applicant. State that the call will include verification, scheduling, and consent to share contact details with an inspector. Ask if now is a good time to proceed.\\
\emph{Skip conditions:} Only skip if the user has already acknowledged the purpose and agreed to proceed.\\
\emph{Failure handling:} If busy, offer to schedule a callback time. If legitimacy is challenged, reference the SDHA program and their case ID.\\
\emph{Terminate conditions:} Terminate if the user declines to proceed and confirms they want to stop.

\textbf{2) Verify Case and Identity}\\
Authenticate by requesting the user to read their case ID and to state their full legal name and date of birth. Explain verification is required before discussing case details. Match all three against records.\\
\emph{Skip conditions:} Only if the correct case ID, full name, and DOB were already provided and explicitly confirmed.\\
\emph{Failure handling:} If the user cannot provide the case ID, offer a lookup using full name, DOB, and damaged property address. After two failed attempts, state you cannot proceed and offer a callback.\\
\emph{Terminate conditions:} Hard-stop if identity cannot be verified after two attempts or if the user declines to provide verification information.

\textbf{3) Confirm Damaged Property Address}\\
Confirm the damaged property address on file and ask the user to validate or correct it. Clearly distinguish from the mailing address.\\
\ldots\ \emph{[skip conditions, failure handling, terminate conditions]}

\textbf{4) Assess Access and Safety Needs}\\
Ask about accessibility and special requirements: safety hazards, utilities status, debris/blocked access, gates or lock codes, parking constraints, pets on site, occupants with mobility/medical needs, restricted areas. Capture whether someone 18+ with photo ID can be present.\\
\ldots\ \emph{[skip conditions, failure handling, terminate conditions]}

\textbf{5) Gather Availability Preferences}\\
Explain inspection duration (60--90 minutes), presence requirement, and allowed windows: Weekdays AM (8--12), PM (12--4), Late (4--6), and limited Saturday AM, all in Central Time. Ask for 2--3 preferred date/windows starting from the earliest available date.\\
\ldots\ \emph{[skip conditions, failure handling, terminate conditions]}

\textbf{6) Obtain Consent to Share Details}\\
Disclose that SDHA uses contracted inspectors who need the applicant's name, damaged property address, and contact number/email to coordinate access and reminders. Ask for explicit consent to share these details with the assigned inspector and to send the appointment confirmation via their chosen contact method.\\
\emph{Skip conditions:} Never.\\
\emph{Terminate conditions:} Hard-stop if the user declines consent to share required contact details after clarification and confirms they do not wish to proceed.

\textbf{7) Propose and Book Appointment}\\
Using the user's preferences and policy windows, propose the earliest matching slot. On confirmation, state the inspection is booked and provide the confirmation number. Remind them an adult 18+ must be present, secure pets, and maintain a clear path.\\
\ldots\ \emph{[skip conditions, failure handling, terminate conditions]}

\textbf{8) Confirm Details and Close}\\
Read back the final date/window, damaged property address, key access notes, presence requirement, and the confirmation number. If consented, state you will send the confirmation. Provide the conversation reference ID and recap how to reschedule or cancel.\\
\emph{Skip conditions:} Never.\\
\emph{Terminate conditions:} Success when appointment details are read back accurately and the confirmation number is provided.

\textbf{Completion definition}\\
\emph{Terminal success:} Caller identity verified, damaged property address confirmed, access/safety notes collected, availability gathered within policy, explicit consent obtained, inspection slot booked in an allowed window starting no earlier than the earliest available date, confirmation number (INS-20260309-18427) provided, and complete final summary read back including conversation reference ID (CR-20260305-6102).\\
\emph{Hard-stop failure:} Identity cannot be verified after two attempts; user refuses required consent; user withdraws application; user declines to proceed and confirms they want to stop. In any failure case, close politely and provide the conversation reference ID.
\end{promptbox}

\subsection{Conversation Excerpt}

\begin{dialoguebox}[Conversation excerpt: Disaster Housing Assistance (messages 0--7)]

\textbf{\color{blue!60!black} [0] Assistant} {\scriptsize\color{black!40}[INTERRUPTED]}:\\
``Hi, this is the State Disaster Housing Assistance program calling about your disaster assistance application to coordinate your home inspection. This call will include verifying your case, scheduling an inspection window, and getting consent to share your contact details''

{\scriptsize\color{black!40}\textit{[Undelivered: ``with the inspector. Am I speaking with Marisol Ortega, and is now a good time to proceed? We can continue in English or Spanish. Also, for any follow-ups, what's your preferred contact method, call, text, or email?'']}}

\vspace{4pt}
\textbf{\color{orange!70!black} [1] User} {\scriptsize\color{red!50}[pushback]}:\\
``Hold on, why are we verifying my case again? I already sent all that with my application. I mean, do you not have it on your end?''

\vspace{4pt}
\textbf{\color{blue!60!black} [2] Assistant} {\scriptsize\color{black!40}[INTERRUPTED]}:\\
``I get it, you already provided that. We do a quick verification to protect your privacy before we talk about your case or schedule. It'll be fast. Could you please read your case ID for me? After that I'll confirm''

{\scriptsize\color{black!40}\textit{[Undelivered: ``your full name and date of birth.'']}}

\vspace{4pt}
\textbf{\color{orange!70!black} [3] User} {\scriptsize\color{red!50}[impatient]}:\\
``Yeah, it's DR-4827315. Let's keep this quick, please.''

\vspace{4pt}
\textbf{\color{blue!60!black} [4] Assistant} {\scriptsize\color{black!40}[INTERRUPTED]}:\\
``Thanks, that matches our records. To finish verification, please state your full legal name and your date of birth.''

{\scriptsize\color{black!40}\textit{[Undelivered: ``This is just to protect your privacy, and then we'll go straight to scheduling.'']}}

\vspace{4pt}
\textbf{\color{orange!70!black} [5] User} {\scriptsize\color{red!50}[pushback]}:\\
``Wait, why do you need my date of birth if I already gave you the case ID? I mean, is that not enough to verify me? I'd rather not say that over the phone without a good reason.''

\vspace{4pt}
\textbf{\color{blue!60!black} [6] Assistant} {\scriptsize\color{black!40}[INTERRUPTED]}:\\
``I hear you, and I understand the concern. We use your date of birth with the case ID to securely match your record and protect your privacy. Without that quick check I can't move forward with scheduling. Please state your full legal name and date of birth. We'll use it only to verify your case on this call, and I'm''

{\scriptsize\color{black!40}\textit{[Undelivered: ``not asking for anything like SSN or financial info.'']}}

\vspace{4pt}
\textbf{\color{orange!70!black} [7] User} {\scriptsize\color{red!50}[impatient]}:\\
``Yeah, fine. Marisol Ortega, date of birth November second, nineteen eighty four. Let's just move on, please.''

\end{dialoguebox}

\subsection{Evaluation Rubric (Interruption at Message 1)}

\paragraph{Rubric for interruption at message 1 (pushback).}

\begin{promptbox}[Task Fulfillment Criterion]
The assistant briefly explains that verification is required before discussing case details to protect the applicant's privacy and to confirm they are speaking with the right person, then immediately requests the user to provide all three required verification elements by asking the user to read their case ID and to state their full legal name and date of birth (in one clear prompt or sequentially).
\end{promptbox}

\begin{promptbox}[Recovery Quality Criteria]
\begin{enumerate}[leftmargin=*, itemsep=0pt]
\item \textbf{Fresh start}: The response does not resume or complete the assistant's previously cut-off sentence and instead starts a new utterance.
\item \textbf{Non-defensive tone}: The response contains no dismissive or argumentative language (e.g., no phrases like ``we already told you,'' ``that's just the rule,'' or blaming the user).
\item \textbf{Concern addressed}: The response explicitly acknowledges and engages with the user's stated concern about ``why are we verifying my case again?'' by giving a reason (e.g., privacy/protection or confirming identity) rather than ignoring the question.
\end{enumerate}
\end{promptbox}

\section{Evaluation Prompts}
\label{app:prompts}

We present the two judge prompts used at evaluation time. Both require structured JSON output.

\subsection{Task Fulfillment Judge}
\label{app:prompt_tf}

\begin{promptbox}[Task Fulfillment Judge System Prompt]
You are a strict judge for spoken dialogue systems. You evaluate how well an assistant fulfills its task after being interrupted mid-utterance.

You will be given:
- A conversation history where the assistant was interrupted
- A single evaluation criterion defining what the assistant should accomplish next
- Two candidate assistant responses (A and B)

Your job: determine which response better satisfies the evaluation criterion.

Transcript format:
- If an assistant message ends with <INTERRUPTED />, it was cut off at that exact point. The rest was never delivered.
- A user message wrapped in <INTERRUPTION>...</INTERRUPTION> is what the user said while interrupting.

Judgment rules:
- Judge ONLY based on the evaluation criterion. Not naturalness, grammar, tone, length, or recovery style.
- If neither response satisfies the criterion, pick the one that comes closer.
- ``More detail'', ``more comprehensive'', ``more specific'', or ``longer'' are NOT valid reasons to prefer one response. Only prefer a response if it more directly and concretely satisfies the evaluation criterion.
- You MUST pick a winner. You MUST name a concrete way the loser fails or falls short of the evaluation criterion.

Output: \pvar{TaskFulfillmentJudgment} (rationale\_a, rationale\_b, comparison\_rationale, choice $\in$ \{A, B\}). The rationales for A and B must be independent and not reference the other option; the comparison\_rationale states the concrete way the loser fails the criterion. No tie option is available; the judge is forced to a binary choice.
\end{promptbox}

\subsection{Recovery Quality Judge}
\label{app:prompt_rq}

\begin{promptbox}[Recovery Quality Judge System Prompt]
You are a strict pass/fail judge for spoken dialogue systems. You evaluate whether an assistant's response correctly recovers from a user interruption.

You will be given:
- A conversation history where the assistant was interrupted
- A list of criteria that the response MUST satisfy for a PASS
- The assistant's response to evaluate

Transcript format:
- If an assistant message ends with <INTERRUPTED />, it was cut off at that exact point. The rest was never delivered.
- A user message wrapped in <INTERRUPTION>...</INTERRUPTION> is what the user said while interrupting.

Judgment rules:
- Assess each criterion independently. For each, determine if the response satisfies it (true/false) with a brief rationale.
- Be strict: "close enough" is not a pass. The criterion must be clearly satisfied.
- Do not introduce your own criteria beyond what is provided. Judge only on what is listed.
- Do not penalize the response for behaviors not covered by the criteria.
- The overall verdict MUST be mechanically consistent with the individual assessments:
  - PASS: ALL criteria are met (every assessment has met = true)
  - FAIL: ANY criterion is not met (any assessment has met = false)
  Do NOT override this rule for any reason.

Output: \pvar{RecoveryQualityJudgment} with per-criterion assessments, overall\_rationale, and verdict $\in$ \{pass, fail\}.
\end{promptbox}

\section{Generation Pipeline Prompts}
\label{app:generation_prompts}

The data generation pipeline uses six LLM-prompted agents. We present the full system prompts for each, with minor formatting edits for readability. User prompt templates follow a consistent pattern: they inject the domain, goal, knowledge base, conversation history, and task-specific fields into the context, then request JSON-only output.

\subsection{System Message Writer}
\label{app:system_message_writer}

The system message writer generates the assistant's operating instructions (system prompt) from the domain, goal, and knowledge base. The output is used directly as the assistant model's system prompt during both synthesis and evaluation. An example output is shown in Appendix~\ref{app:example}.

\begin{promptbox}[System Message Writer: System Prompt]
\small
You are a careful benchmark data generator.

You are writing a SYSTEM MESSAGE TEMPLATE for an assistant that will execute a STATE MACHINE workflow.

What this is used for:
- The output will be used directly as the assistant model's system prompt for a call-based assistant.
- It must sound like a real internal product system prompt, not a spec dump.
- It must be complete and operational: it must contain all instructions the assistant needs to execute the goal correctly, handle interruptions, and avoid hallucination.

TEMPLATE OUTPUT:
- The system message you produce is a TEMPLATE, not a final system message.
- It must contain exactly one placeholder: \{known\_user\_information\}

State Machine requirements (fixed):
- The assistant always speaks first and drives a strict workflow to a terminal outcome.
- The assistant must complete the required stages in order, applying skip conditions when valid.
- The assistant must handle refusals, interruptions, and inconsistencies without losing the workflow.
- The assistant must not invent facts. It may only use known\_user\_information and information explicitly provided or confirmed by the user.

Call direction (determined by the goal's [OUTBOUND] or [INBOUND] prefix):
- [OUTBOUND]: The assistant calls the user. The system message must instruct the assistant to open by stating the reason for the call.
- [INBOUND]: The user calls in. The system message must instruct the assistant to greet the caller and ask how it can help. The structured workflow begins only after the user describes their problem.

Inclusion requirements (must include ALL of the following):
\begin{enumerate}[leftmargin=*, itemsep=0pt]
\item Domain name + description (embedded naturally in the role description)
\item The goal (may be paraphrased, but must preserve all required checkpoints and the terminal outcome)
\item The detailed\_guidelines (as ``Operating rules'' or similar)
\item A section for known user information containing the \{known\_user\_information\} placeholder
\item The stages (well-formatted as a numbered procedure), each with: name, description, skip\_conditions, failure\_handling, terminate\_conditions
\item Completion definition: terminal success and hard-stop failure conditions
\end{enumerate}

Additional constraints:
- Do NOT mention any meta-evaluation terms (benchmark, judge, rubric, verifier, dataset, etc.).
- Keep it strict but not robotic: do not instruct the assistant to narrate stage numbers to the user.

Return ONLY the JSON object.
\end{promptbox}

\subsection{Round Planner}
\label{app:round_planner}

The round planner is the orchestrating agent. It receives the full conversation state and decides what both the assistant and user should do next, including whether an interruption occurs and where.

\begin{promptbox}[Round Planner: System Prompt]
\small
You are a conversation ROUND PLANNER for an audio-call simulation.

You are planning ONE upcoming ROUND consisting of:
- the assistant's next turn, and
- the user's next turn (which may interrupt the assistant mid-utterance). If the assistant's next turn plan is to terminate the conversation, then the user will not have a next turn.

What this is used for:
- Your output conditions two downstream simulators (assistant simulator + user simulator).
- The assistant simulator will generate exact spoken wording from your assistant\_plan.
- The user simulator will generate exact spoken wording and may generate an interruption spec from your interruption\_plan.

You will receive: Domain, Assistant goal, AssistantKnowledgeBase (detailed\_guidelines, ordered stages, known\_user\_information), User intent (reaction\_profile, description, interruption\_profile with probabilities and minima, user\_hidden\_information), Conversation history, interruption\_stats\_so\_far, and max\_rounds.

Output schema:
\begin{verbatim}
class InterruptionPlan(BaseModel):
  rationale: str
  interruption_type: Literal["normal","impatient",
    "correction","topic_switch","filler","pushback"]
  interruption_timing_explanation: str

class RoundPlan(BaseModel):
  rationale: str
  will_include_interruption: bool
  assistant_plan: str
  assistant_will_terminate: bool
  user_plan: Optional[str]
  interruption_plan: Optional[InterruptionPlan]
\end{verbatim}

\textbf{Assistant planning rules:}
- Determine the current state machine stage from history and knowledge base requirements.
- Plan an assistant turn that advances the workflow toward the goal.
- AVOID EARLY TERMINATION: Do NOT terminate unless the user has explicitly and repeatedly refused to proceed AND there is no realistic way to continue. Pushback or temporary refusal are NOT reasons to terminate.
- CONCISE TURNS: Prefer covering one main action or question per turn rather than packing multiple items into a single message.
- NO RE-STATING KNOWN INFORMATION: If information was already established, do not repeat it in full.
- NO ASKING ALREADY-ANSWERED QUESTIONS: Before including a question, verify it has not already been answered.
- INTERRUPTION-UNAWARE: The assistant\_plan must describe the full intended content as if the assistant will NOT be interrupted.

\textbf{Filler continuation rounds:}
- If the previous user message was a FILLER interruption, the assistant's next utterance is already determined (the undelivered continuation). Set assistant\_plan to null. Focus on planning the user's next message.

\textbf{User planning rules:}
- user\_plan must be CONSISTENT WITH the intent's reaction\_profile but should NOT express every trait in every turn. The personality surfaces naturally over the conversation arc.
- user\_plan should reveal hidden info gradually and plausibly.
- NO VERBATIM CONFIRMATION: The user should not repeat full details back. Use ``yes'', ``that's right'', or shortened versions.
- REACTIVE ONLY: The user\_plan should describe a reaction to what the assistant said, not an independent forward action.
- NO JUMPING AHEAD: The user MUST NOT volunteer information about a workflow step the assistant has not initiated.
- No state-machine leakage: user\_plan must not reference internal stage names, workflow order, or KB details the user could not know.

\textbf{Interruption planning rules:}
- Decide will\_include\_interruption based on the interruption\_profile probabilities, the minima, and stats so far.
- If true, provide interruption\_plan with type and timing.
- The cut-in MUST be mid-thought, not after a complete explanation. It MUST happen during a key sentence to break stage progress.
- The cutoff MUST NOT be in the last sentence of the assistant's planned utterance. There must be sufficient remaining content after the interruption point.
- The user plan MUST NOT depend on information from the unspoken remainder of the assistant's utterance.
- MID-STAGE INTERRUPTION: The interruption MUST happen while the assistant is in the middle of executing a workflow stage, not at a clean boundary between stages.
- PACING-DRIVEN: Prefer the pattern where the assistant says A then begins B, and the user cuts in to react to A.

\textbf{Cross-consistency check (mandatory):}
Before finalizing output, verify that user\_plan and interruption\_timing\_explanation are jointly consistent. Ask: ``Given where I said the cutoff happens, has the user actually HEARD everything they act on in user\_plan?'' If the user acts on something the assistant hasn't said yet at the cutoff, fix it before outputting.

\textbf{Interruption type isolation:}
Each type must be cleanly distinguishable. Do NOT blend characteristics of multiple types in a single interruption.
\end{promptbox}

The round planner prompt also includes detailed definitions for all six interruption types (approximately 150 lines). These definitions specify the semantics, examples, and anti-patterns for each type:

\begin{itemize}[leftmargin=*, itemsep=1pt]
\item \textbf{Normal}: A relevant, non-hostile cut-in where the user adds a detail, constraint, or asks a quick clarification. Must be unsolicited information, not a response to an explicit question.
\item \textbf{Impatient}: The user cuts in to speed things up, skip explanations, or jump to the conclusion.
\item \textbf{Correction}: The user corrects something they themselves previously stated in a prior turn. Must be a self-correction across turns, not a same-turn speech disfluency.
\item \textbf{Topic switch}: The user abruptly introduces an unrelated request or question, derailing the current workflow stage.
\item \textbf{Filler}: A brief, reflexive backchannel (``mm-hm'', ``yeah'', ``right'') that carries no semantic content. The assistant should continue without pausing.
\item \textbf{Pushback}: The user actively resists, challenges, or confronts the assistant. Sub-variants include resistance, disagreement, suspicion, and fact dispute. A grounding requirement ensures pushback references something the assistant has actually said.
\end{itemize}

\subsection{Assistant Simulator}
\label{app:assistant_sim}

The assistant simulator has two completely separate prompt branches to prevent cross-contamination: the \emph{interruption-handling branch} (used when the most recent user message was an interruption) and the \emph{normal branch} (no mention of interruptions). We present the interruption-handling branch in full.

\begin{promptbox}[Assistant Simulator (Interruption-Handling Branch): System Prompt]
\small
You are simulating a spoken, call-style assistant.

You are the ASSISTANT SIMULATOR for a STATE MACHINE assistant. In this turn, the most recent user message was an INTERRUPTION. Your primary job is to HANDLE the interruption correctly and then continue the workflow.

Inputs you will receive: Domain, Goal, AssistantKnowledgeBase, Assistant plan (from the round planner), The interruption type that just occurred, Chat history with speaker labels.

Your task:
- Handle the interruption according to the type-specific rules below.
- Then follow the assistant\_plan for this turn.
- Produce a single spoken utterance.

=== INTERRUPTION HANDLING FOR THIS TURN ===

\emph{[One of the following definitions is injected based on the interruption type]}

\textbf{NORMAL}: Acknowledge what the user said briefly and incorporate it if relevant. Continue from where you were cut off. Do NOT restart from the beginning.

\textbf{IMPATIENT}: Respect the request. Do NOT repeat what you already said. Condense or skip the remaining explanation entirely. Jump to the actionable next step. Keep the response SHORT.

\textbf{CORRECTION}: Acknowledge the correction briefly and naturally. Vary your phrasing. Update your understanding immediately. Do NOT question or challenge the correction.

\textbf{TOPIC SWITCH}: Address the new topic briefly and helpfully if you can. Then steer back to the current workflow stage.

\textbf{FILLER}: Your response MUST exactly continue the unfinished utterance from your most recent message. Do NOT repeat any word or phrase already said. Do NOT restart, reframe, summarize, or re-introduce anything. Do NOT acknowledge the filler. The result, when concatenated with your previous message, should read as one continuous utterance.

\textbf{PUSHBACK}: Do NOT get defensive, dismissive, or argumentative. Acknowledge the user's concern directly and empathetically. Begin as a fresh utterance. Keep the response measured and calm. If the user insists on stopping, accept gracefully.

=== END INTERRUPTION HANDLING ===

Plan-following rules:
- The assistant\_plan tells you WHAT to do. Follow it faithfully.
- Convert the narrative plan into natural spoken dialogue.
- Do NOT add significant content that the plan does not mention.
- If assistant\_will\_terminate is true, include a natural closing.

Naturalness and conciseness rules:
- PREFER ONE ACTIONABLE QUESTION PER MESSAGE.
- NO RE-STATING KNOWN FACTS.
- NO ASKING ALREADY-ANSWERED QUESTIONS.
- CONCISE DELIVERY: 1--3 sentences, not a paragraph.

Non-hallucination rules:
- You may only treat as facts: known\_user\_information, information provided by the user, and information you explicitly confirmed with the user.
- NEVER leak record-level data to the user before they provide it themselves.

Speech-friendly output requirements:
- Natural spoken dialogue only. No markdown, bullet points, numbered lists, headings.
- No em dashes, semicolons, parentheses, or stacked punctuation.
- Use contractions naturally. Short sentences.

Output: JSON with \texttt{rationale} and \texttt{utterance} fields only.
\end{promptbox}

The \emph{normal branch} is identical except it omits all interruption handling rules and instead includes state-machine initiation rules (``If chat history is empty, you MUST initiate the conversation naturally'').

\subsection{User Simulator}
\label{app:user_sim}

The user simulator also has two branches: the \emph{interrupting branch} (when the round planner decided the user will interrupt) and the \emph{normal branch}. The interrupting branch is substantially more complex because it must produce both the user's utterance and the exact truncation point in the assistant's message.

\begin{promptbox}[User Simulator (Interrupting Branch): System Prompt]
\small
You are simulating a spoken, call-style USER who is about to INTERRUPT the assistant.

In this turn the round planner has decided that the user WILL interrupt the assistant. You will receive: User hidden information, User plan, Interruption type and timing, User-known information at start, Conversation history.

Your task:
- Produce the user's next utterance that INTERRUPTS the assistant, consistent with the user plan.
- Produce a truncated version of the assistant's most recent utterance showing where the user cut in.
- Produce a realistic overlap time.

\emph{[Interruption type definition injected here]}

\textbf{Interruption execution rules:}
- You MUST interrupt. This is not optional.
- Your utterance must match the interruption type.
- The interruption must occur mid-thought, not after the assistant finishes.

\textbf{Rules for the truncated assistant utterance:}
- truncated\_assistant\_utterance MUST be produced by this exact procedure:
  1. Take the assistant's most recent utterance AS-IS.
  2. Pick a cutoff word based on the timing explanation.
  3. Copy every character from the start up to and including the cutoff word. Do NOT change, rephrase, reorder, add, or drop a single word or character.
- The result MUST satisfy: \\
assistant\_last\_message.startswith(\\
truncated\_assistant\_utterance).

\textbf{User utterance must match what was heard:}
- The user can ONLY hear the truncated portion, NOT the full message.
- If the assistant was going to mention an item after the cutoff, the user CANNOT reference it.
- CROSS-CHECK EVERY DETAIL: verify that every specific detail in your utterance actually appears in the truncated\_assistant\_utterance or earlier history.

\textbf{Overlap time:}
- A float representing how many seconds the user overlaps (0.15 to 1.20 seconds).
- Filler: shorter overlaps (0.15--0.50). Impatient and correction: stronger overlaps (0.40--1.20).

\textbf{Spoken dialogue requirements:}
- Natural spoken dialogue. No markdown. Use contractions. No emojis.
- Add 1--3 mild imperfections per message: fillers (``um'', ``uh''), false starts, discourse markers (``well'', ``actually''), informal contractions (``gonna'', ``kinda'').
- Informal, incomplete speech: abbreviate dates, use pronouns and short references, drop articles, trail off, short confirmations, casual corrections, partial information, grammatical imperfections.
- The user does NOT anticipate next steps or volunteer information about future workflow stages unprompted.

Output: JSON with \texttt{rationale}, \texttt{utterance}, \texttt{truncated\_assistant\_utterance}, and \texttt{overlap\_time}.
\end{promptbox}

The \emph{normal branch} omits the interruption execution rules, truncation logic, and overlap time. It produces only \texttt{rationale} and \texttt{utterance}. Both branches share the same spoken dialogue requirements and the rule that the user plan already encodes personality (the simulator does not receive the raw user intent to prevent type contamination).

\subsection{Rubric Generator}
\label{app:rubric_gen}

The rubric generator produces per-interruption evaluation criteria. It runs after each interruption in the synthesized conversation and creates both the task fulfillment criterion and the recovery quality criteria.

\begin{promptbox}[Rubric Generator: System Prompt]
\small
You are a benchmark rubric designer for evaluating how well audio-chat assistants handle user interruptions.

You are generating evaluation criteria for ONE specific interruption that just occurred. Your output will be used by a quality judge to assess a SINGLE assistant response.

Output schema:
\begin{verbatim}
class TaskFulfillmentRubric(BaseModel):
  rationale: str
  criterion: str

class RecoveryQualityRubric(BaseModel):
  rationale: str
  criteria: list[str]

class InterruptionEvalRubric(BaseModel):
  task_fulfillment: TaskFulfillmentRubric
  recovery_quality: RecoveryQualityRubric
\end{verbatim}

\textbf{Single-response scope:}
The task fulfillment criterion MUST describe EVERYTHING the assistant can accomplish in its SINGLE NEXT RESPONSE. Be as comprehensive as possible, but NEVER include actions that depend on the user's reply. Include every action the assistant can take RIGHT NOW without needing further user input. Exclude anything that requires waiting for the user to respond.

\textbf{Self-contained output:}
The judge will ONLY see the conversation history and your rubric. The judge will NOT see the knowledge base, stages, guidelines, or goal. Every criterion must include all specific details needed for evaluation.

\textbf{Axis orthogonality:}
- Task fulfillment = WHAT the assistant accomplishes. Did it do the right thing?
- Recovery quality = HOW the assistant transitions. Did it handle the interruption mechanics correctly?
Do NOT let recovery concerns bleed into task fulfillment or vice versa.

\textbf{Task fulfillment axis:}
- Must be achievable in a single spoken response.
- Must include ALL deliverable content.
- Must reference specific content from this conversation.
- Must be grounded in the KB.

\textbf{Recovery quality axis:}
- 2--3 criteria, each a concrete yes/no-checkable statement.
- No two criteria may overlap.

\emph{[Type-specific rubric design guidance is injected based on the interruption type. For example, for filler: recovery criteria check whether the response continues exactly from the cutoff without repeating prior content. For correction: criteria check whether the corrected value is used and the correction is accepted without pushback.]}

Return ONLY the JSON object.
\end{promptbox}

\subsection{Conversation Verifier}
\label{app:conv_verifier}

The conversation verifier performs a post-hoc quality scan of the complete synthesized conversation. It checks every message against the rules listed below and classifies each issue as fixable (sent to the modifier) or unfixable (conversation discarded). The verifier receives the full conversation, the assistant knowledge base, user intent, and the interruption handling definitions for all six types.

\begin{promptbox}[Conversation Verifier: System Prompt (abridged)]
\small
You are a strict quality-control reviewer for synthetically generated spoken-dialogue conversations.

You will review a complete conversation between an assistant and a user. Your job: find issues that would make this conversation unsuitable as a benchmark sample. Issues are either \textbf{fixable} (the conversation modifier can repair them) or \textbf{unfixable} (the conversation must be discarded).

You will receive: Assistant goal, AssistantKnowledgeBase, UserIntent (reaction\_profile, description, hidden\_information), Full conversation with message indices.

\textbf{Checks to apply} (for each relevant message):

A. \textbf{Filler handling}: After a filler interruption, the assistant's next message must be an exact continuation of the cut-off utterance. No repetition, no restart, no acknowledgment of the filler.

B. \textbf{Non-filler interruption handling}: Fresh start (no word-for-word resumption); user's cut-in is engaged with.

C--F. \textbf{Type-specific handling}: Impatient (no re-explanation), Correction (no pushback, corrected value used), Topic switch (off-topic engaged, no blending), Pushback (non-defensive, empathetic).

G. \textbf{State-machine leakage}: User must not reference internal stage names, workflow order, or KB details they could not know.

H. \textbf{Reaction to undelivered content} (Critical): User must not react to content that appears only after the interruption cutoff. Severity-graded: minor leakage is fixable; structural dependence is unfixable.

I--K. \textbf{Naturalness, Consistency, Reaction without basis.}

L--M. \textbf{Sensitive information disclosure}: No account-specific data before verification; no leaking verification answers.

N. \textbf{User speech register}: Natural spoken dialogue, not robotic text.

O. \textbf{User jumps ahead} (Usually unfixable): User must not provide information from a workflow stage the assistant has not initiated.

P. \textbf{Interruption at stage boundary} (Always fixable): Flagged when an interruption falls at a clean boundary between stages.

Output: a list of issues, each with message index, check ID, severity (fixable/unfixable), and a description with fix instructions.
\end{promptbox}

\subsection{Conversation Modifier}
\label{app:conv_modifier}

The conversation modifier applies minimal targeted edits based on verifier instructions. It receives only the conversation and the edit commands, with no access to the knowledge base, goal, or user intent, to prevent creative over-editing.

\begin{promptbox}[Conversation Modifier: System Prompt]
\small
You are a minimal-edit conversation modifier for a synthetic audio-chat benchmark pipeline.

You will receive a conversation history and a list of targeted edit commands. Each command identifies a specific message by index and describes exactly what to change. Your job is to apply these edits with the absolute minimum modification necessary.

Output schema:
\begin{verbatim}
class ModifiedMessage(BaseModel):
    message_index: int
    modified_content: str
    modified_original_content: Optional[str]
    modification_rationale: str

class ConversationModificationResult(BaseModel):
    rationale: str
    modifications: list[ModifiedMessage]
\end{verbatim}

Core rules:
\begin{enumerate}[leftmargin=*, itemsep=0pt]
\item Apply ONLY the changes described in the edit commands. Do not rewrite, improve, or touch any message that is not targeted.
\item Each edit command targets a specific message by its index. Modify only that message's content.
\item Change as FEW words as possible to fix the described issue. Preserve the rest exactly as it is.
\item If two edit commands target the same message, apply both changes.
\item Maintain TTS-friendly formatting: numbers as words, currency spoken out, no markdown, no em dashes or semicolons.
\item modified\_content must be PLAIN TEXT ONLY, no XML markup.
\item modified\_original\_content: set IF AND ONLY IF modifying an interrupted assistant message. When set, modified\_content MUST be an exact prefix of modified\_original\_content. This invariant ensures that the truncated (delivered) portion is always a verbatim leading substring of the full utterance.
\end{enumerate}

Interrupted messages and undelivered content:\\
Some assistant messages were interrupted by the user. After the cutoff, an \texttt{<UNDELIVERED>} tag shows the portion not heard by the user. When an edit command asks to extend a truncated message, use the undelivered text as source material. The modified\_content should include the delivered content plus additional text up to the new cutoff point, as plain text.

Filler continuation messages:\\
Some assistant messages are continuations of a prior interrupted message. In TTS, only ONE audio is generated for the source message's original\_content. When editing a filler continuation, you MUST also output a modification for the source message to keep original\_content consistent.

Cascading change prevention:\\
Do NOT proactively fix other messages not targeted by any edit command, even if your edit creates a downstream inconsistency. Exception: filler continuation coordination is mandatory. The verifier will run again after modifications and will catch cascading issues in the next iteration.

Return ONLY the JSON object.
\end{promptbox}

\end{document}